\pdfoutput=1
\documentclass[12pt, draftclsnofoot, onecolumn]{IEEEtran}
\IEEEoverridecommandlockouts
\usepackage{amsmath,graphicx,epstopdf,amssymb,amsthm}
\usepackage{cite} 
\usepackage{hyperref}
\usepackage{epstopdf} 
\usepackage{makecell}
\usepackage{amsmath,bm}    
\usepackage{verbatim} 
\usepackage{array}
\usepackage[utf8]{inputenc}     
\usepackage[english]{babel}   
\usepackage{caption}    
\usepackage{enumitem} 
\usepackage[ruled,vlined,linesnumbered]{algorithm2e}
\DeclareCaptionLabelFormat{lc}{\MakeLowercase{#1}~#2}
\usepackage[dvipsnames]{xcolor}
 
\newenvironment{skproof}{\noindent\textit{Sketch of  Proof:}}{\hfill$\blacksquare$}

\newtheorem{theorem}{Theorem}
\newtheorem{lemma}{Lemma} 
\newtheorem{fact}{Fact}
\newtheorem{definition}{Definition}
\newtheorem{proposition}{Proposition}
\newtheorem{corollary}{Corollary}

\newtheorem{assumption}{Assumption}

\allowdisplaybreaks
\usepackage[protrusion=true,expansion=true]{microtype}
\pdfoutput=1

\usepackage[font=small]{caption}

\usepackage{float}
\usepackage{mathtools}

\DeclarePairedDelimiter\floor{\lfloor}{\rfloor}

\newcommand{\nm}[1]{{\color{red} {\bf [NM: #1]}}}
\newcommand{\add}[1]{\textcolor{red}{#1}}

\usepackage{cancel}
   
\usepackage{soul,xcolor}
\DeclareMathOperator*{\argmin}{arg\,min}

\def\BibTeX{{\rm B\kern-.05em{\sc i\kern-.025em b}\kern-.08em
    T\kern-.1667em\lower.7ex\hbox{E}\kern-.125emX}}
\begin{document}
\setulcolor{red}
\setul{red}{2pt}
\setstcolor{red}
\title{Delay-Aware Hierarchical Federated Learning}





\author{Frank Po-Chen Lin,~\IEEEmembership{Student Member,~IEEE},  Seyyedali~Hosseinalipour,~\IEEEmembership{Member,~IEEE}, Nicol\`o Michelusi, \IEEEmembership{Senior~Member,~IEEE}, and Christopher G. Brinton,~\IEEEmembership{Senior~Member,~IEEE}
\thanks{F. Lin and C. Brinton are with the School of Electrical and Computer Engineering, Purdue University, IN, USA. e-mail: \{lin1183,cgb\}@purdue.edu. Brinton and Lin acknowledge support from ONR grants N000142212305 and N000142112472.}
\thanks{S. Hosseinalipour is with the Department of Electrical Engineering, University at Buffalo, NY, USA. e-mail: alipour@buffalo.edu.}
\thanks{N. Michelusi is with the School of Electrical, Computer and Energy Engineering, Arizona State University, AZ, USA. e-mail: nicolo.michelusi@asu.edu. Part of his research has been funded by NSF under grant CNS-2129015.}
\thanks{A condensed version of this paper was presented at IEEE Globecom 2020~\cite{frank2020delay}.}}
\maketitle

\begin{abstract}
Federated learning has gained popularity as a means of training models distributed across the
wireless edge.
The paper introduces delay-aware hierarchical federated learning ({\tt DFL}) to improve the efficiency of distributed machine learning (ML) model training by accounting for communication delays between edge and cloud. Different from traditional federated learning, {\tt DFL} leverages multiple stochastic gradient descent iterations on local datasets within each global aggregation period and intermittently aggregates model parameters through edge servers in local subnetworks. During global synchronization, the cloud server consolidates local models with the outdated global model using a local-global combiner, thus preserving crucial elements of both, enhancing learning efficiency under the presence of delay. A set of conditions is obtained to achieve the sub-linear convergence rate of $\mathcal O(1/k)$ for strongly convex and smooth loss functions. Based on these findings, an adaptive control algorithm is developed for {\tt DFL}, implementing policies to mitigate energy consumption and communication latency while aiming for sublinear convergenc. Numerical evaluations show {\tt DFL}'s superior performance in terms of faster global model convergence, reduced resource consumption, and robustness against communication delays compared to existing FL algorithms. In summary, this proposed method offers improved efficiency and results when dealing with both convex and non-convex loss functions.
\end{abstract} 




Machine Learning (ML) has become a popular tool for carrying out a variety of practical applications in   computer vision, speech recognition, natural language processing, and robotic control~\cite{wu2017squeezedet,AI_jordan,goldberg2017neural}. Conventionally, ML model training for these applications is conducted in a centralized manner, where
data from different sources is collected and processed in a single server/datacenter.
Nevertheless, in many applications, data used for model training is generated/gathered at the modern Internet-of-Things (IoT) devices located at the edge of the network (e.g., autonomous vehicles, mobile phones, and wearable devices)~\cite{CVN}, which makes centralized model training impractical. In fact, transferring the massive amount of data collected from the IoT devices to a central location imposes high latency and power/resource consumption~\cite{Chiang}, which is not desired especially in real-time applications~\cite{wu2017squeezedet,hard2018federated}. 
Also, privacy concerns related to transmitting private data across the network have progressively advocated for storing data locally and shifting the ML model training to the network edge where the data gets collected. This has led to an emerging area of distributed ML over the network edge, which exploits the distributed computing power of IoT devices to realize an intelligent edge/fog~\cite{hosseinalipour2020federated,park2019wireless}.

Federated learning (FL)~\cite{mcmahan2017communication}, a popular distributed ML framework, trains an ML model via engaging edge devices in collaborative model training while keeping their data locally.
It does so by executing three steps repeatedly: (i) \textit{local model training} at each edge device using its local dataset, (ii) \textit{aggregation of the local models} to a global model by a server, and (iii) \textit{synchronization of the local models} with the newly obtained global model.
Upon being implemented over a real-world network edge, FL faces many challenges and design problems given the heterogeneities existing at the wireless edge \cite{lim2020federated,hosseinalipour2020federated}. (i) \textit{Extreme Data Heterogeneity:}
the local datasets of the devices may exhibit significant heterogeneity, making them non-independent and identically distributed (non-i.i.d.), causing the locally trained models in each edge device to be significantly biased towards the local dataset~\cite{hosseinalipour2020federated}. 
(ii) \textit{Delay of Model Aggregations:} Variable distances and quality of communications between the edge devices and the point of aggregation (e.g., a cloud server) may result in considerable model aggregation and synchronization delays, making naive synchronization of the received global model with the local model impractical~\cite{frank2020delay}. 
(iii) \textit{Hierarchical Architecture of Network Edge:} Large-scale edge networks do not admit the conventional FL network architecture.
In particular, edge devices are often not directly connected to a cloud server, instead, they are connected to the edge servers, which facilitate the communication to the cloud server~\cite{wang2021HFL}.
In this paper, we are motivated to address the above three challenges. In particular, we consider model training under a
metric of data heterogeneity extended from the current art, which can capture extreme cases of data diversity across the devices. Also, we explicitly account for the delay in model aggregation and introduce a linear local-global model combining scheme. This scheme retains essential elements of both the outdated global model and the current local model, thereby improving the overall learning efficiency. Our methodology can augment existing studies by incorporating our strategy with those already established. Finally, we consider model training over a realistic hierarchical network edge architecture.

\textbf{Summary of Contributions}\label{subsec:contrib}:\\
Our contributions in this work can be summarized as follows:
\begin{itemize}[leftmargin=5mm]
\item We propose \textit{delay-aware federated learning} ({\tt DFL}), a novel methodology for improving distributed ML model training efficiency by accounting for the round-trip delay between edge and cloud. 
{\tt DFL} accounts for the effects of delays by introducing a local-global model combiner scheme during global synchronization, which conserve vital aspects of both the stale global model and the current local model, thereby enhancing overall learning efficiency.

\item  We theoretically investigate the convergence behavior of {\tt DFL} under a generalized data heterogeneity metric. Our convergence analysis introduces new techniques, such as coupled dynamic systems, to the current art in hierarchical FL~\cite{Feng2022HFL,Lim2021HFL,Xu2022HFL,Luo2020HFEL,wang2021HFL,Mhaisen2022HFL}. We obtain a set of conditions to achieve the sub-linear convergence rate of $\mathcal{O}(1/k)$ for strongly convex and smooth loss functions while mitigating the communication delay, which resembles the convergence rate of stochastic gradient descent in centralized model training without delay.

\item Leveraging the convergence characteristics, we introduce an adaptive control algorithm for {\tt DFL}, which targets a joint optimization of communication energy, latency, and ML performance while preserving a sub-linear convergence rate. This involves solving a non-convex integer programming problem, adapting (i) the global aggregation interval with the cloud, (ii) the local aggregation interval with the edge servers, (iii) the learning rate, and (iv) the combiner weight over time.

\item Our numerical evaluations demonstrate the effectiveness of {\tt DFL} in terms of convergence speed and resource consumption under various network settings, under both convex and non-convex loss functions. We observe that {\tt DFL} achieves (i) faster convergence of the global model, (ii) reduced resource consumption, and (iii) improved robustness against communication delays compared to existing FL algorithms.
\end{itemize}
The structure of this paper is as follows: Section~\ref{sec:RW} delves into works related to this study, specifically addressing the challenges mentioned in Federated Learning (FL). Subsequently, Section~\ref{sec:tthf} outlines the system model and details the machine learning methodology implemented in {\tt DFL}. A theoretical analysis of {\tt DFL}'s convergence behavior is provided in Section~\ref{sec:convAnalysis}, followed by a discussion on our adaptive control algorithm designed to optimize controllable parameters within {\tt DFL} in Section~\ref{Sec:controlAlg}. Lastly, Section~\ref{sec:experiments} showcases numerical experiments conducted to assess the performance of {\tt DFL}.
For brevity, proofs are provided as sketches with full versions in the appendices.

\section{Related Work}\label{sec:RW}
We categorize the related work to this study with respect to the three aforementioned challenges of FL. For a comprehensive survey of works on FL, we refer the interested reader to~\cite{Kairouz}.

\textbf{Non-i.i.d. Data Across the Devices.}
Non-i.i.d data across the devices has been shown to significantly reduce the performance of FL due to local model bias \cite{mcmahan2017communication}. To counteract this effect, \cite{wang2019adaptive,Sun2021adaptive} tunes the frequency of global model aggregations, \cite{tu2020network,wang2021device} conduct data transfer across the devices to reduce the heterogeneity of data, and~\cite{Tran2019cap,chen2019joint,yang2020energy} conduct efficient resource allocation under non-i.i.d. data. However, most of these works rely on a simple modeling of non-i.i.d. data across the devices which cannot be generalized to real-world settings.
Recently, we introduced a new metric of data heterogeneity in \cite{lin2021timescale} that extends the current art and is able to capture extreme non-i.i.d. data across the devices. However, in \cite{lin2021timescale}, this metric is exploited in a completely different framework, where devices conduct local device-to-device (D2D) communications for model consensus, as compared to this works. In this work, we aim to conduct convergence analysis of model training under this general data heterogeneity metric, while taking into account for communication delay and hierarchical network architecture.


\textbf{Model Training under Delay Considerations.} Upstream/downstream communication between the edge and server can
often lead to non-negligible transmission delays in FL. 
Delay of model training in FL has been modeled and considered in several recent works \cite{Tran2019cap,PSL2022federated,Shi2021latency,sam2022latency,Zhao2022layency,Gao2021latency}. These works model the delay with respect to the channel conditions between the devices and the server and the devices' local computation power. The delay is often aimed to be minimized in these works to have the fastest model training scheme. However, none of these works aim to \textit{mitigate} the impact of delay via intelligent synchronization of the received global model and the local model at the devices. In this work, we study this under-explored topic via proposing a linear local-global model combiner.

\textbf{Hierarchical Federated Learning.}
Several studies have aimed to propose new system architectures to improve the scalability of ML model training in FL~\cite{liu2020client,Lim2022EI,wang2021HFL,lin2021timescale,Zhao2020HFL}. Specifically, hierarchical federated learning has been proposed as a realistic system model, which allows edge devices to first transmit their local model parameters to edge servers for intermediate aggregation before edge servers conduct global aggregation through a cloud server, which reduces the frequency of performing resource intensive global aggregations~\cite{Feng2022HFL,Lim2021HFL,Xu2022HFL,Luo2020HFEL,wang2021HFL,Mhaisen2022HFL}. This architecture has been investigated through the lens of resource allocation \cite{Xu2022HFL,Feng2022HFL}, and edge association~\cite{Luo2020HFEL,wang2021HFL,Lim2021HFL,Mhaisen2022HFL}. Although these works provide valuable design insights, they mostly consider a simplistic data heterogeneity metric and do not take into account for smart local synchronization of the local models with the global model. As compared to these works, we introduce a new ML convergence analysis scheme that takes into account for a generalized data heterogeneity metric under a new linear local-global model combiner scheme. Drawing from the convergence properties, we devise a control algorithm for {\tt DFL}, which incorporates strategies aimed at mitigating energy consumption and communication latency.
\section{System Model and Machine Learning Methodology}
\label{sec:tthf}
\noindent In this section, we first describe our hierarchical edge network system model (Sec.~\ref{subsec:syst1}) and formalize the ML task under consideration (Sec.~\ref{subsec:syst2}). Then, we develop our delay-aware federated learning procedure, {\tt DFL} (Sec.~\ref{subsec:syst3}). A list of acronyms and symbols used is provided in Table~\ref{table:res} and~\ref{table:notation}.

\begin{figure*}[t]
  \centering
    \includegraphics[width=0.96\textwidth]{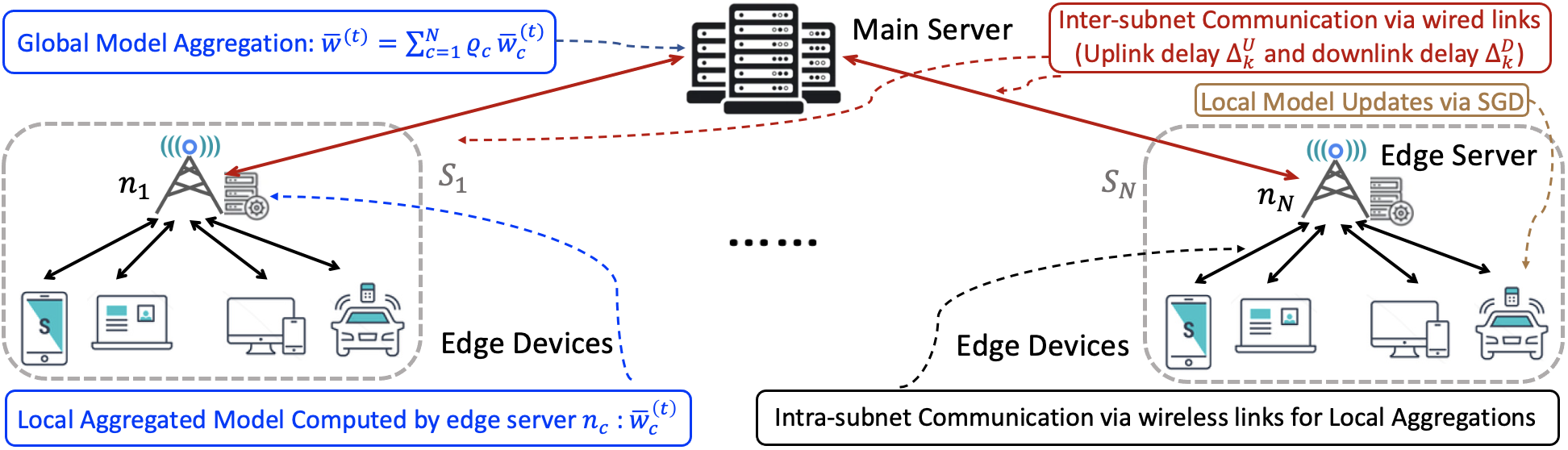}
     \caption{
     The three-layer hierarchical federated learning network architecture consists of edge devices forming local subnet topologies based on their proximity to the local aggregator.}
     \label{fig2}
     \vspace{-0.7em}
\end{figure*}

\subsection{Edge Network System Model}
\label{subsec:syst1}
We consider ML model learning over the hierarchical network architecture depicted in Fig.~\ref{fig2}. The network consists of three layers of nodes. At the top of the hierarchy, 
a main server (e.g., a cloud server) is in charge of global model aggregations. In the middle of the hierarchy, there are $N$ edge servers (e.g., coexisting with cellular base stations), collected via the set $\mathcal{N}=\{n_1,\cdots,n_N\}$, which act as local model aggregators. Finally, at the bottom of the hierarchy, there are $I$ edge devices represented via the set $\mathcal{I}=\{1,\cdots,I\}$ in charge of local model training. 
The edge devices are organized into subnets (subnetworks), with a one-to-one mapping between the subnets and the edge servers, providing them uplink/downlink connectivity. In particular, 
 the devices
are partitioned into $N$ sets $\{\mathcal S_c\}_{c=1}^{N}$, where set $\mathcal{S}_c$ is associated with edge server $n_c \in \mathcal{N}$ for uplink/downlink model transmissions, with
$\mathcal{S}_c \cap \mathcal{S}_{c'} =\emptyset~\forall c\neq c'$ and $\cup_{c=1}^{N} \mathcal{S}_{c}=\mathcal{I}$. 
Each subnet $\mathcal{S}_c$ consists of $s_c = |\mathcal{S}_c|$ edge devices, where $\sum_{c=1}^{N}s_c=I$.\footnote{This model is capable of being adapted to time-varying topologies, for instance,  mobile devices switching subnet. However, we omit this for notational simplicity.}

\begin{table}[h]
  \centering
  \caption{List of acronyms}
  \label{table:res}
  \begin{tabular}{|c|c|c|c|}
    \hline
    \textbf{Acronym} & \textbf{Definition} & \textbf{Acronym} & \textbf{Definition} \\
    \hline
    ML & Machine Learning & FL & Federated Learning \\
    \hline
    DFL & Delay-Aware Federated Learning & SGD & Stochastic Gradient Descent \\
    \hline
    SVM & Support Vector Machine & CNN & Convolutional Neural Network \\
    \hline
  \end{tabular}
\end{table}

\begin{table}[h]
  \centering
  \caption{List of key symbols}
  \label{table:notation}
  \begin{tabular}{|c|m{6.3cm}|c|m{6.3cm}|}
    \hline
    \textbf{Symbol} & \textbf{Description} & \textbf{Symbol} & \textbf{Description} \\
    \hline
    $\mathcal N$ & Set of all edge servers & $N$ & Number of all edge servers \\
    \hline
    $\mathcal I$ & Set of all edge devices & I & Number of all edge devices \\
    \hline
    $\mathcal S_c$ & Set of edge devices in subnet $c$ & $s_c$ & Number of edge devices in subnet $c$ \\
    \hline
    $n_c$ & Index of local model aggregator (edge server) & $\ell(\cdot)$ & Loss function associated with each datapoint \\
    \hline
    $\mathcal D_i$ & Dataset of edge devices $i$ & $D_i$ & Number of data points in edge device $i$ \\
    \hline
    $F_i(\cdot)$ & Local loss function for device $i$ & $\bar{F}_c(\cdot)$ & Subnet loss function for subnet $c$ \\
    \hline
    $F(\cdot)$ & Global loss function & $\mathbf w^*$ & Optimal global model \\
    \hline
    $\delta, \zeta$ & Inter-gradient diversity parameters across subnets & $\delta_c,\zeta_c$ & Intra-gradient diversity parameters in subnet $c$ \\
    \hline
    $t$ & Time index of local model training iteration & $k$ & Index of local model training interval\\
    \hline
    $\mathcal T_k$ & $k$-th local model training interval & $\tau_k$ & Length of the $k$-th local model training interval \\
    \hline
    $\mathcal T_{k,c}^\mathsf{L}$ & Set of local aggregation instances in the $k$-th local model training interval & $\widehat{\mathbf g}_{i}^{(t)}$ & Local stochastic gradient at device $i$ and time $t$ \\
    \hline
    $\mathbf w_i^{(t)}$ & Local model at device $i$ and time $t$ & $\widetilde{\mathbf w}_i^{(t)}$ & Tentative Local model at device $i$ and time $t$ \\
    \hline
    $\bar{\mathbf{w}}_{c}^{(t)}$ & Instantaneous aggregated local model across devices $i\in\mathcal S_c$ & $\bar{\mathbf w}^{(t)}$ & Global model at time $t$ \\
    \hline
    $\Theta_c^{(t)}$ & Indicator of local aggregation for subnet $c$ & $\alpha_k$ & Combiner weight for the linear local-global combiner \\
    \hline
    $\Delta^\mathsf{U}_k$ & Upstream delay between edge device and main server & $\Delta^\mathsf{D}_k$ & Downstream delay between edge device and main server \\
    \hline
  \end{tabular}
\end{table}

\subsection{Machine Learning Model} \label{subsec:syst2}
We assume that each edge device $i\in\mathcal{I}$ owns a dataset $\mathcal{D}_i$ with $D_i=|\mathcal{D}_i|$ data points. In general, the $\mathcal{D}_i$'s are non-i.i.d. (non-independent and identically distributed), i.e., there exists statistical dataset diversity across the devices. Each data point $(\mathbf x,y)\in\mathcal D_i$, $\forall i$, consists of an $m$-dimensional feature vector $\mathbf x\in\mathbb R^m$ and a label $y\in \mathbb{R}$.
We let $\ell(\mathbf x,y;\mathbf w)$ denote the \textit{loss} associated with 
data point $(\mathbf x,y)$ under \textit{ML model parameter vector} $\mathbf w \in \mathbb{R}^M$, where $M$ denotes the dimension of the model. The loss quantifies the precision of the ML model with respect to the underlying ML task; for example, in linear regression, $\ell(\mathbf x,y;\mathbf w) = \frac{1}{2}(y-\mathbf w^\top\mathbf x)^2$. The \textit{local loss function} at device $i$ is defined as
\begin{align}\label{eq:1}
    F_i(\mathbf w)=\frac{1}{D_i}\sum\limits_{(\mathbf x,y)\in\mathcal D_i}
    \ell(\mathbf x,y;\mathbf w).
\end{align}

We subsequently define the \textit{subnet loss function} for each $\mathcal{S}_c$ as the average loss of devices in the subnet, i.e.,
\begin{align}\label{eq:c}
    \bar F_c(\mathbf w)=\sum\limits_{i\in\mathcal{S}_c} \rho_{i,c} F_i(\mathbf w),
\end{align}
where $\rho_{i,c} = D_i/\sum_{j\in\mathcal S_c}D_j$ is the weight associated with edge device $i\in \mathcal{S}_c$ relative to its respective subnet. 
The \textit{global loss function} is defined as the average loss across the subnets
\begin{align} \label{eq:2}
    F(\mathbf w)=\sum\limits_{c=1}^{N} \varrho_c \bar F_c(\mathbf w),
\end{align}
where $\varrho_c = \sum_{i\in\mathcal S_c}D_i/\sum_{j\in\mathcal I}D_j$ is the weight associated with subnet $\mathcal{S}_c$ relative to the network.

The goal of ML model training is to find the optimal global model parameter vector $\mathbf w^* \in \mathbb{R}^M$: 
\begin{align}
    \mathbf w^* = \mathop{\argmin_{\mathbf w \in \mathbb{R}^M} }F(\mathbf w).
\end{align}

In the following, we introduce some  assumptions regarding the above-defined loss functions that are commonly employed in distributed ML literature~\cite{haddadpour2019convergence,wang2019adaptive,chen2019joint,friedlander2012hybrid}. These assumptions further imply the existence and uniqueness of $\mathbf w^*$. 
\begin{assumption}[Loss Functions Characteristics]\label{Assump:SmoothStrong}
\label{beta}
Henceforth, the following assumptions are made:
\begin{itemize}[leftmargin=3mm]
\item  \textbf{Strong convexity}:
 Each local loss function $F_i$ is $\mu$-strongly convex~$\forall i\in\mathcal I$, i.e.,
\begin{equation} \label{eq:11_mu} 
    F_i(\mathbf w_1) \geq  F_i(\mathbf w_2)+\nabla F_i(\mathbf w_2)^\top(\mathbf w_1-\mathbf w_2)
    +\frac{\mu}{2}\Big\Vert\mathbf w_1-\mathbf w_2\Big\Vert^2 \hspace{-1.2mm},~\hspace{-.5mm}\forall { \mathbf w_1,\mathbf w_2} \in  \mathbb{R}^M.
\end{equation} where $\Vert\cdot\Vert$ refers to the Euclidean norm.
    \item  \textbf{Smoothness:} Each local loss function $F_i$ is $\beta$-smooth $\forall i\in\mathcal{I}$, i.e., 
    \begin{align} \label{eq:11_beta}
\Big\Vert \nabla F_i(\mathbf w_1)-\nabla F_i(\mathbf w_2)\Big\Vert \leq & \beta\Big\Vert \mathbf w_1-\mathbf w_2 \Big\Vert, ~\forall \mathbf w_1, \mathbf w_2 \in \mathbb{R}^M,
 \end{align}
where $\beta>\mu$. These assumptions imply the strong convexity and $\beta$-smoothness of $\bar{F}_c$ and $F$. \footnote{Throughout, $\Vert \cdot \Vert$ is  used to denote 2-norm of the vectors.}

\end{itemize}

\end{assumption}
The above assumptions are leveraged in our theoretical analysis in Sec.~\ref{sec:convAnalysis}, and subsequently to develop the control algorithm in Sec.~\ref{Sec:controlAlg}. Our experiments in Sec.~\ref{sec:experiments} demonstrate the effectiveness of our methodology, even for non-convex loss functions (e.g., neural networks).

We next introduce measures of gradient diversity to quantify the statistical heterogeneity across local datasets. Different from our existing work~\cite{lin2021timescale}, we consider this both across and within subnets, which will be important to our analysis:



\begin{definition}[Inter-Subnet Gradient Diversity]\label{gradDiv}
The inter-subnet gradient diversity across the device subnets is measured via two non-negative constants $\delta,\zeta$ that satisfy 
\begin{align} \label{eq:11}
    \left\Vert\nabla\bar F_c(\mathbf w)-\nabla F(\mathbf w)\right\Vert
    \leq \delta+ \zeta \Vert\mathbf w-\mathbf w^*\Vert,~\forall c, \mathbf w.
\end{align}
\end{definition} 
\begin{definition}[Intra-Subnet Gradient Diversity]\label{gradDiv_c}
The intra-subnet gradient diversity across the devices belonging to subnet $\mathcal{S}_c$ is measured via non-negative constants $\delta_c,\zeta_c$ that satisfy
\begin{align} \label{eq:clust_div}
    \left\Vert\nabla F_i(\mathbf w)-\nabla\bar F_c(\mathbf w)\right\Vert
    \leq \delta_c+\zeta_c\Vert\mathbf w-\mathbf w^*\Vert,~\forall i\in\mathcal{S}_c,  ~\forall c, \mathbf w.
\end{align}
\end{definition}
The definitions of gradient diversity are used to measure the degree of heterogeneity across the local datasets of devices in federated learning settings, which may be non-i.i.d., and can impact the convergence performance. A high value of gradient diversity implies a large dissimilarity between the local datasets and the global data distributions.
We further define ratios $\omega=\frac{\zeta}{2\beta}\leq 1$ and $\omega_c=\frac{\zeta_c}{2\beta}\leq 1$, which will be important in our analysis. The above definitions are obtained using the $\beta$-smoothness of the loss functions. Specifically, for~\eqref{eq:11},
\begin{align}\label{eq:motivNewDiv}
    &\Vert\nabla\bar F_c(\mathbf w)-\nabla F(\mathbf w)\Vert
    = \Vert\nabla\bar F_c(\mathbf w)-\nabla\bar F_c(\mathbf w^*)+\nabla\bar F_c(\mathbf w^*)+\underbrace{\nabla F(\mathbf w^*)}_{=0}-\nabla F(\mathbf w)\Vert
    \nonumber \\&
    \leq \Vert\nabla\bar F_c(\mathbf w)-\nabla\bar F_c(\mathbf w^*)\Vert+\Vert\nabla\bar F_c(\mathbf w^*)\Vert + \Vert\nabla F(\mathbf w)-\nabla F(\mathbf w^*)\Vert
    \leq \delta+2\beta\Vert\mathbf w-\mathbf w^*\Vert,
\end{align}
where $\Vert\nabla\bar F_c(\mathbf w^*)\Vert\leq\delta$. The bound in~\eqref{eq:motivNewDiv} demonstrates the fact that by applying the smoothness property of the functions $\bar F_c(\cdot)$ and $F(\cdot)$ along with imposing an upper bounded $\delta$ on the subnet gradients at the optimum, the definition of gradient diversity in~\eqref{eq:clust_div} is a general expression for~\eqref{eq:motivNewDiv},
which in turn results in~\eqref{eq:11} for $\frac{\zeta}{2\beta}\leq 1$. 
Using the same steps~\eqref{eq:clust_div} can be obtained, as we discuss in~\cite{lin2021timescale}.
These gradient diversity metrics extend 
the conventional definition of gradient diversity used in literature, e.g., as in \cite{wang2019adaptive}, which is a special case of~\eqref{eq:11} and~\eqref{eq:clust_div} with $\zeta=\zeta_c=0$, $\forall c$. Given that in FL settings, the local dataset of the devices may be extremely non-i.i.d., very large values of $\delta$ and $\delta_c$ in~\eqref{eq:11} and~\eqref{eq:clust_div} are required when $\zeta=\zeta_c=0$, which will, in turn, make the convergence bounds very loose and ineffective in describing the system behavior.
The addition of terms with coefficients $\zeta, \zeta_c$ in~\eqref{eq:11} and~\eqref{eq:clust_div} will lead to bounded
values for $\delta$ and $\delta_c$, especially in the beginning of training when $\Vert \mathbf w-\mathbf w^*\Vert$ can take an arbitrary large value. At the same time, the addition of $\zeta, \zeta_c$ in~\eqref{eq:11},~\eqref{eq:clust_div} forms a coupling between the gradient diversity and the optimality gap $\Vert \mathbf w-\mathbf w^*\Vert$. As we will see, this coupling makes our convergence analysis unique from the current art and rather non-trivial.

\subsection{{\tt DFL}: Delay-Aware Federated Learning} \label{subsec:syst3}
\subsubsection{Overview and Rationale}
\begin{figure*}[t]
\includegraphics[width=\textwidth]{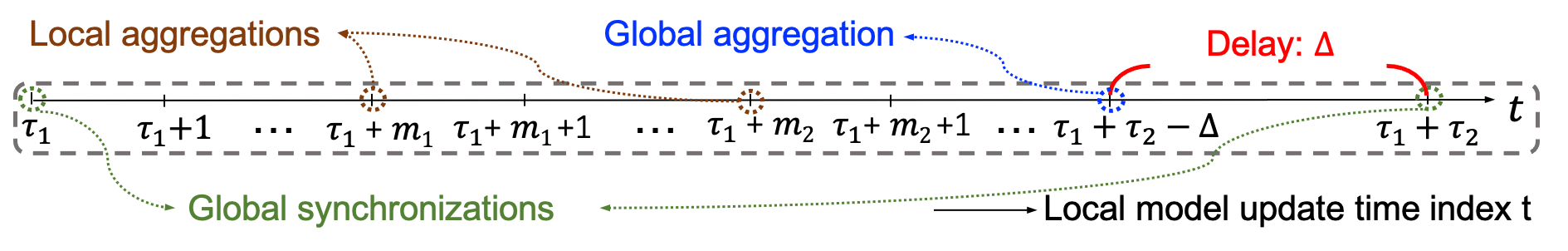}
\centering
\caption{Illustration of the timescale in {\tt DFL}. The time index $t$ represents local model updates through SGD, local aggregation, as well as global aggregations and synchronizations.}
\label{fig:twoTimeScale}
\vspace{-5mm}
\end{figure*}

In the context of {\tt DFL}, the process of training the ML model comprises of a sequence of local model training intervals that occur between two successive \textit{global synchronizations}, which are subsequently followed from \textit{global aggregations}. During each interval, the edge devices carry out stochastic gradient descent (SGD) iterations based on their local data to optimize their local loss functions $F_i(\cdot),~\forall i$. Additionally, the edge server conducts aperiodic \textit{local aggregations}\footnote{Note that in Fig.~\ref{fig:twoTimeScale}, the parameters $m_1$ and $m_2$ may not be equal. In fact, {\tt DFL} allows for aperiodic local aggregations during each local model training interval, the frequency of which will be later tuned in our control algorithms to mitigate network resource consumption.} to synchronize the local model parameters with the edge devices within its corresponding subnet. Aggregation refers to computing an \textit{aggregated model} by taking the weighted average of local models from edge devices. In contrast, synchronization refers to updating local models at edge devices using the aggregated model obtained after aggregation. On one hand, local aggregations are conducted within subnets through device-to-edge communication links, resulting in small-scale communication delays because edge servers are typically physically closer to the edge devices. On the other hand, global aggregations occur across the entire network through edge-to-cloud communication links, which typically have non-negligible propagation delays in the range of hundreds of milliseconds to several seconds, depending on network bandwidth~\cite{Lin}. This delay can significantly degrade learning performance in FL because the main server is often far from the edge devices. In this paper, we neglect the delay of local aggregations within subnets and focus on the global aggregation delay, assuming device-to-edge links have a much shorter range than edge-to-cloud links. Details of the {\tt DFL} procedure and the modeling of communication delay will be described in the following.


\subsubsection{{\tt DFL} Model Training Procedure}\label{subsec:proc} In the following, we provide a high-level description of our methodology, which will be later formalized in Sec.~\ref{subsec:form2}.
We consider a slotted-time representation of a network where each edge device conducts a \textit{local model training iteration} via SGD at each time index $t=0,1,\cdots,T$. The time duration of $0$ to $T$ is partitioned into multiple \textit{local model training intervals} indexed by $k=0,1,\cdots,K-1$, each capturing time interval $\mathcal{T}_k = \{t_{k} + 1,...,t_{k+1}\}\subset \{0,1,\cdots,T\}$. The model training procedure at the devices starts with the server broadcasting the initial global model $\bar{\mathbf w}^{(0)}$ to all the devices at $t_0=0$, proceeding through a series of global aggregations.  Local model training intervals occur in between two consecutive global model synchronizations at time $t_k$ and $t_{k+1}$, during which the local models of devices are updated via SGD. 
At each instance of global aggregation, the models of the devices are collected (i.e., they engage in uplink transmissions) and arrive at the server with a certain delay. The server will then aggregate the received models into a global model and broadcast it (i.e., engage in downlink transmission) to the devices, which involves another delay. In parallel, devices proceed with local updates before the global model arrives. Upon reception of the global model, devices synchronize their local models in a manner accounting for both uplink and downlink delays. The relationships between the timescales are  depicted in Fig.~\ref{fig:twoTimeScale}.

The length of the $k$th local model training interval is denoted by $\tau_k = |\mathcal{T}_k|$. We have $t_k = \sum_{\ell=0}^{k-1}\tau_\ell$, $\forall k$, where $t_0 = 0$ and $t_K=T$. This allows
for varying the length of local model training intervals across global synchronizations (i.e., possibly $\tau_{k} \neq \tau_{k'}$ for $k \neq k'$).
During the $k$th local model training interval, we define $\mathcal{T}^\mathsf{L}_{k,c} \subset \mathcal{T}_k$ as the set of time instances when the edge server pulls the models of devices in subnet $\mathcal{S}_c$ (i.e., devices engage in uplink transmissions) and performs local aggregations of their models. Local aggregation synchronizes the local models within the same subnet by computing their weighted average.
We next formalize the above procedure.

\subsubsection{Formalizing {\tt DFL}}\label{subsec:form2}
Next, we formalize the local training, local model aggregations, and global model aggregation and synchronization. Through this process, we will also introduce new notions used for orchestrating
the formulation of the system. 
\begin{definition}[Conditional Expectation]\label{def:cond_E}
$\mathbb{E}_t(\cdot)$ represents the conditional expectation conditioned on $\mathcal F_t$, where $\mathcal F_t$ denotes the $\sigma$-algebra generated by all random sampling up to but excluding, $t$.
\end{definition}

\textbf{Local SGD iterations}: 
At $t \in \mathcal{T}_k$, device $i$ randomly samples a mini-batch $\xi_i^{(t)}$ of datapoints from its local dataset $\mathcal D_i$, and uses it to calculate the unbiased \textit{stochastic gradient estimate} using its previous local model $\mathbf w_i^{(t)}$ as
\vspace{-2mm}
\begin{align}\label{eq:SGD} 
    \widehat{\mathbf g}_{i}^{(t)}=\frac{1}{\vert\xi_i^{(t)}\vert}\sum\limits_{(\mathbf x,y)\in\xi_i^{(t)}}
    \nabla\ell(\mathbf x,y;\mathbf w_i^{(t)}).
\end{align}

\textbf{Tentative Local Model}: Using the gradient estimate $\widehat{\mathbf g}_{i}^{(t-1)}$, each device computes its \textit{tentative local model} $ {\widetilde{\mathbf{w}}}_i^{(t)}$ as
\vspace{-3.5mm}
\begin{align} \label{8}
    {\widetilde{\mathbf{w}}}_i^{(t)} = 
           \mathbf w_i^{(t-1)}-\eta_{k} \widehat{\mathbf g}_{i}^{(t-1)},~t\in\mathcal T_k, 
\end{align}
where $\eta_{k} > 0$ denotes the step size. We denote this tentative because it is an intermediate calculation prior to any potential aggregation. Specifically, 
based on ${\widetilde{\mathbf{w}}}_i^{(t)}$, the \textit{updated local model} $\mathbf{w}_i^{(t)}$ is computed either through setting it to ${\widetilde{\mathbf{w}}}_i^{(t)}$ or through a local aggregation, discussed next.

\textbf{Updated Local Model}: 
At time $t$, if subnet $\mathcal{S}_c$ does not conduct a local aggregation, i.e., $t\in \mathcal T_k\setminus \mathcal T^\mathsf{L}_{k,c}$, the final updated local model is obtained based on the the conventional rule ${{\mathbf{w}}}_i^{(t)} = {\widetilde{\mathbf{w}}}_i^{(t)}$ from \eqref{8}. Otherwise, i.e., if $t\in \mathcal T^\mathsf{L}_{k,c}$, then each device $i$ in subnet $\mathcal{S}_c$ transmits its tentative updated local model $\tilde{\mathbf w}_i^{(t)}$ to edge server 
$n_c$ which computes the \textit{instantaneous aggregated local model} as follows:
\vspace{-2mm}
\begin{equation}
    \label{eq:local_agg}
    \bar{\mathbf w}_c^{(t)} = \sum_{i\in\mathcal{S}_c} \rho_{i,c} \tilde{\mathbf w}_i^{(t)}.
\end{equation}
The edge server $n_c$ then broadcasts $\bar{\mathbf w}_c^{(t)}$ across its subnet. The devices subsequently obtain their final updated models with local aggregation as
    ${{\mathbf{w}}}_i^{(t)} =\bar{\mathbf w}_c^{(t)}, i\in\mathcal{S}_c$.

Based on the above described procedure, we can obtain the following general local model update rule at each device $i\in\mathcal S_c$:
\vspace{-2mm}
\begin{align} \label{eq:w_i-gen}
    \mathbf w_i^{(t)} = (1-\Theta_c^{(t)})\widetilde{\mathbf{w}}_i^{(t)}+\Theta_c^{(t)}\bar{\mathbf w}_c^{(t)},~\forall t \in \mathcal T_k,
\end{align}
\vspace{-0mm}
where ${\Theta_c^{(t)}}$ is the indicator of local model aggregation, defined as
\vspace{-0mm}
\begin{align} \label{eq:the}
        {\Theta_c^{(t)}} = 
        \begin{cases}
            0  ,& t\in \mathcal T_k\setminus \mathcal T^\mathsf{L}_{k,c} \\
            1 ,& t\in \mathcal T^\mathsf{L}_{k,c}.
        \end{cases}
\end{align} 

\textbf{Communication Delay and Global Model Aggregations}: 
In this work, we explicitly consider the communication delay from the edge to the cloud during model training. During global aggregation $k$, the upstream communication delay of local models from devices to the main server is denoted as $\Delta^\mathsf{U}_k$, while the downstream delay of the global model to edge devices is denoted as $\Delta^\mathsf{D}_k$. We then introduce the \textit{round-trip communication delay} to describe the duration from the moment edge devices transmit their local models to the edge servers (which subsequently forward them to the main server) until the devices finalize their local model synchronization using the received global model, expressed as $\Delta_k=\Delta^\mathsf{U}_k+\Delta^\mathsf{D}_k$, assuming $0\leq \Delta_k \leq \tau_k-1,~\forall k$. Both uplink $\Delta^\mathsf{U}_k$ and downlink delay $\Delta^\mathsf{D}_k$ are quantified in terms of the \textit{number} of local SGD update rounds that edge devices can perform during this period.

In conducting global aggregations, to account for the round-trip delay, devices send their local models to the edge servers $\Delta_k$ time instances prior to the completion of each local model training interval $\mathcal{T}_k$, i.e., $t=t_{k+1}-\Delta_k \in \mathcal{T}_k$. Concurrently, these devices carry out an additional $\Delta_k$ local updates using local model updates before they receive the updated global model.
We assume that $\Delta_k$ can be reasonably estimated, e.g., from round-trip delays observed in recent time periods\footnote{In this study, we utilize the most recently observed delay, measured during the last local model training interval, as an approximation for the forthcoming round-trip delay, drawing inspiration from TCP analysis where round-trip times remain fairly stable for tens of seconds \cite{hao2002RTT}.} 


The edge servers then forward the locally aggregated models $\bar{\mathbf w}_{c}^{(t)}$ to the main server with the delay of $\Delta^{\mathsf{U}}_k$. The main server builds the global model based on the \textit{stale} local models as

\begin{align} \label{15}
    \bar{\mathbf w}^{(t)} &= 
          \sum\limits_{c=1}^N \varrho_c \bar{\mathbf w}_{c}^{(t)}, \;\; t=t_{k+1}-\Delta_k, k=0,1,...,
\end{align}
where $\bar{\mathbf w}^{(t)}=\sum\limits_{c=1}^N \varrho_c \bar{\mathbf w}_{c}^{(t)}$ is the global average of local models for all $t$ and $\varrho_c$ is defined in~\eqref{eq:2}. Note that the computation of the global model $\bar{\mathbf w}^{(t_{k+1}-\Delta_k)}$ is performed by the main server at $t=t_{k+1}-\Delta^{\mathsf{D}}_k$. After the computation at the main server, the global model $\bar{\mathbf w}^{(t_{k+1}-\Delta_k)}$ is then broadcast and received at the devices with the delay of $\Delta^{\mathsf{D}}_k$. The devices then synchronize their local models at time $t_{k+1}$ via a linear local-global model combiner, as follows.

\begin{algorithm}[t]
\scriptsize 
\SetAlgoLined
\caption{Delay-aware federated learning {\tt DFL} with set control parameters.} \label{DFL}
\KwIn{Length of training $T$, number of global aggregations $K$, local aggregation instances $\mathcal T^\mathsf{L}_{k,c}, c=1,...,N$, length of local model training intervals $\tau_k,~\forall k$, combiner weights $\alpha_k,~\forall k$, learning rates $\eta_k,~\forall k$, minibatch sizes $\vert\xi_i^{(t)}\vert,~ i\in\mathcal{I},~\forall t$} 
\KwOut{Final global model $\bar{\mathbf w}^{(T)}$}
 // Initialization by the server \\
 Initialize $\bar{\mathbf w}^{(0)}$ and broadcast it across edge devices, resulting in $\mathbf w_i^{(0)}=\bar{\mathbf w}^{(0)}$, $i\in\mathcal I$.
 \\
 \For{$k=0:K-1$}{
     \For{$t=t_{k}+1:t_{k+1}$}{
        \For(// Procedure at each subnet $\mathcal{S}_c$){$c=1:N$}
        {
         Local SGD update 
         with:
          $\widetilde{\mathbf w}_i^{(t)} = 
          \mathbf w_i^{(t-1)}-\eta_{t-1} \widehat{\mathbf g}_{i}^{(t-1)}$\; 
        \uIf{$t\in\mathcal T^\mathsf{L}_{k,c}$}{
        Local model aggregation with:
        $\bar{\mathbf w}_c^{(t)} = \sum_{i\in\mathcal{S}_c} \rho_{i,c} \tilde{\mathbf w}_i^{(t)}$ and $\mathbf w_i^{(t)}=\bar{\mathbf w}_c^{(t)}$\;
        }
        \Else{
        $\mathbf{w}_i^{(t)} = \widetilde{\mathbf{w}}_i^{(t)}$.
        }
      }
      \uIf{$t=t_{k+1}-\Delta_k$}{
      \For(// Procedure at each subnet $\mathcal{S}_c$){$c=1:N$}{
      Local model aggregation: 
      $\bar{\mathbf w}_c^{(t)} = \sum_{i\in\mathcal{S}_c} \rho_{i,c} \tilde{\mathbf w}_i^{(t)}$ and uplink transmission.\\
      }
          }
      \uElseIf( // Procedure at main server){$t=t_{k+1}-\Delta_k^{\mathsf{D}}$}{
      Global model aggregation: $\bar{\mathbf w}^{(t_{k+1}-\Delta_k)}= \sum\limits_{c=1}^N \varrho_c \bar{\mathbf w}_{c}^{(t_{k+1}-\Delta_k)}$ and downlink broadcast.\\}
      \ElseIf( // Procedure at each cluster $\mathcal{S}_c$){$t=t_{k+1}$ and $t\neq T$}{
     Local-Global model combiner:
     $\mathbf w_i^{(t)} = (1-\alpha_k) \bar{\mathbf{w}}^{(t-\Delta_k)}
    +\alpha_k\left((1-\Theta_c^{(t)})\widetilde{\mathbf{w}}_i^{(t)}+\Theta_c^{(t)}\sum_{j\in\mathcal{S}_c} \rho_{j,c} \widetilde{\mathbf w}_j^{(t)}\right)$.\\
      }
 }
}
\end{algorithm}

\textbf{Linear Local-Global Model Combiner (Global Synchronization)}: The conventional rule for updating the local model in FL is to synchronize based on the global model, i.e., $\mathbf w_i^{(t_{k})} = \bar{\mathbf w}^{(t_{k+1}-\Delta_k)}$. However, in our setting, the devices have conducted $\Delta_k$ more local updates before receiving the global model, which this standard synchronization would effectively neglect, resulting in synchronizing the local model with stale global model $\bar{\mathbf w}^{(t_{k+1}-\Delta_k)}$.
To address this, we propose a linear global-local model combiner scheme in which each edge device $i$ updates its local model based on the received global model at time $t_{k+1}$ as
\begin{align} \label{eq:aggr_alpha}
    \hspace{-0.156in}
    \mathbf w_i^{(t_{k+1})} \hspace{-0.05in}&= \hspace{-0.02in} (1-\alpha_k) \bar{\mathbf{w}}^{(t_{k+1}-\Delta_k)}
    +\alpha_k\left((1-\Theta_c^{(t_{k+1})})\widetilde{\mathbf{w}}_i^{(t_{k+1})}+\Theta_c^{(t_{k+1})}\sum_{j\in\mathcal{S}_c} \rho_{j,c} \widetilde{\mathbf w}_j^{(t_{k+1})}\right)\hspace{-0.05in}, \hspace{-0.08in}~\forall i \in \mathcal{I},
\end{align}
where $\alpha_k \in [0,1)$ is the \textit{combiner weight} employed in update iteration $k$, with $\alpha_k=0$ corresponding to the conventional synchronization rule. Devices will then commence their local SGD iterations over $\mathcal{T}_{k+1}$ initialized based on~\eqref{eq:aggr_alpha}.
Intuitively, $\alpha_k$ should be carefully tuned to compensate for the tradeoff between the \textit{staleness} of the global model and the potential for local model \textit{overfitting} to each device's dataset. In particular, when we have a larger delay $\Delta_k$, $\alpha_k$ 
is expected to be larger since the global model received will be based on more outdated local models. In Sec.~\ref{Sec:controlAlg}, we will develop a control algorithm (i.e., Algorithm~\ref{GT}) to determine $\tau_k$ and $\alpha_k$ given round-trip delay $\Delta_k$ for each local model training interval.  

The pseudocode for the {\tt DFL} algorithm with preset control parameters is given in Algorithm~\ref{DFL}. In Section~\ref{Sec:controlAlg}, we present the corresponding control algorithm (Algorithm~\ref{GT}) that tunes the algorithm parameters to achieve a sublinear convergence rate based on the bound derived in Sec.~\ref{sec:convAnalysis}, while mitigating network costs.

\section{Convergence Analysis of {\tt DFL}} \label{sec:convAnalysis}
\noindent In this section, we aim to provide a theoretical analysis of the convergence behavior of the global deployed model under {\tt DFL}. To facilitate this analysis, 
we adopt an approach similar to~\cite{err_free2022}. We break down the errors
between the local models and the global optimum into: 1) the error between the local models and \emph{virtual} models following noise-free dynamics; and 2) the errors between the latter and the global optimum. To this end,
we define a virtual subnet \textit{noise-free variable}
(i.e., it considers the full-batch gradient of $\bar{F}_c(\cdot)$, thereby neglecting the SGD noise and intra-subnet diversity) that remains constant throughout the entire training period as
\begin{align} \label{eq:errFree_c}
    \bar{\mathbf v}_c^{(t+1)}=\bar{\mathbf v}_c^{(t)}
        -\eta_{k}\nabla \bar{F}_c(\bar{\mathbf v}_c^{(t)}),~\forall t\in\mathcal{T}_k\setminus{\{t_k\}}.
\end{align}
The \textit{subnet noise-free variable at global synchronization} is defined as 
\begin{align}\label{eq:EFGS}
    \bar{\mathbf v}_c^{(t_{k+1})}
    =& (1-\alpha_k)\bar{\mathbf v}^{(t_{k+1}-\Delta_k)}
    +\alpha_k\widetilde{\mathbf v}_c^{(t_{k+1})},
\end{align}
where $\widetilde{\mathbf v}_c^{(t_{k+1})}$ is the noise-free variable right before global synchronization, as opposed to $\mathbf v_c^{(t_{k+1})}$ defined immediately after global synchronization. Similarly, we define the \textit{virtual global noise-free variable} as  
\begin{equation} \label{eq:v_m}
    \bar{\mathbf v}^{(t+1)}=\sum\limits_{d=1}^N\varrho_{d}\bar{\mathbf v}_d^{(t+1)},~\forall t\in\mathcal{T}_k.
\end{equation}
To simplify the presentation of the convergence analysis, we assume $\alpha_k=\alpha$ (i.e., the weighing coefficient in \eqref{eq:EFGS}) $\tau_k=\tau$ (i.e., the length of local model training interval) and $\Delta_k=\Delta$ (i.e., the delay), are constants throughout the training, for all $k$.
This matches the design of the control algorithm presented in Algorithm~\ref{GT} (Sec.~\ref{Sec:controlAlg}), where an optimization formulation is formulated at the beginning of each global synchronization to optimize the performance metrics of our interest for the remaining of ML model training time and determine $\tau_k$ and $\alpha_k$ for each aggregation given the edge-to-cloud communication delay $\Delta_k$. Although $\alpha$, $\tau$, and $\Delta$ are assumed to be fixed in the convergence analysis, we will later use the analysis results to obtain instantaneous $\alpha_k$ for {\tt DFL}.



\subsection{Intermediate Quantities and Results}
We make the following assumptions and define three quantities used throughout the analysis.
\begin{assumption}[SGD Noise Characteristics] \label{assump:SGD_noise}
    Let ${\mathbf n}_{i}^{(t)}=\widehat{\mathbf g}_{i}^{(t)}-\nabla F_i(\mathbf w_{i}^{(t)})$
    $\forall i,t$ denote the noise of the estimated gradient through the SGD process for device $i$ at time $t$. The conditional expectation based on time $t$ is $\mathbb{E}_t[{\mathbf n}_{i}^{(t)}]=\bm{0}$ with an upper bound $\sigma^2$ on the variance of the noises, such that $\exists \sigma>0: \mathbb{E}_t[\Vert{\mathbf n}_{i}^{(t)}\Vert^2]\leq \sigma^2, \forall i,t$. 
\end{assumption}
\begin{assumption}[Subnet Deviation Noise] \label{assump:sub_err}
    We assume that $\Theta_c^{(t)}$ is chosen such that the following noise within a subnet is upper bounded by $\phi^2$
    \begin{align}
        \sum\limits_{c=1}^N\varrho_c(1-\Theta_c^{(t)})(2\delta_c^2+4\omega_c^2\beta^2\Vert\bar{\mathbf v}_c^{(t)}-\mathbf w^*\Vert^2)\leq\phi^2,
    \end{align}
    where $\phi>0$ is a constant. 
\end{assumption}
We will ensure the enforcement of the above assumption through the control algorithm in Sec.~\ref{Sec:controlAlg}.
We now define the following set of error terms.
\subsubsection{Subnet Deviation Error}
We define the \textit{subnet deviation error} as follows:  
\begin{align} \label{def:eps_c}
   e_1^{(t+1)}\triangleq\mathbb E\Big[\sum\limits_{c=1}^N\varrho_c\sum_{j\in\mathcal S_c}\rho_{j,c}\Vert\mathbf w_j^{(t+1)} - \bar{\mathbf v}_c^{(t+1)}\Vert^2\Big]^{1/2}.
\end{align} 
$e_1^{(t)}$ captures the average deviation error of local models of devices within a subnet from the subnet noise-free variable.
\subsubsection{Expected Model Dispersion and Optimality Gap of the Noise-Free Variable}
We next define 
\begin{align}\label{eq:defA}
e_2^{(t)} = \sum\limits_{c=1}^N\varrho_{c}\mathbb E[\Vert\bar{\mathbf v}_c^{(t)}-\bar{\mathbf v}^{(t)}\Vert]
\end{align}
as the \textit{expected model dispersion} of the noise-free variable with $\bar{\mathbf v}_c^{(t)}$ and $\bar{\mathbf v}^{(t)}$ defined in~\eqref{eq:errFree_c} and~\eqref{eq:v_m} respectively. $e_2^{(t)}$ measures the degree to which the 
subnet noise-free variable deviates from the global noise-free variable during the local model training interval. In addition, we define
\begin{align}\label{eq:defB}
    e_3^{(t)}=\mathbb E[\Vert\bar{\mathbf v}^{(t)}-\mathbf w^*\Vert]
\end{align}
as the \textit{expected optimality gap of the Noise-Free variable}. $e_3^{(t)}$ measures the degree to which the global noise-free variable deviates from the optimum during the local model training interval.

Obtaining a general upper bound on $e_2^{(t)}$ and $e_3^{(t)}$ is non-trivial due to the coupling between the gradient diversity and the model parameters imposed by~\eqref{eq:11} and~\eqref{eq:clust_div}. For an appropriate choice of step size in~\eqref{8}, we upper bound these quantities by analyzing coupled variable systems. Upper bounds of $e_1^{(t)}$, $e_2^{(t)}$ and $e_3^{(t)}$ are illustrated in the following Lemma and Proposition. Using these bounds, we later obtain the convergence bounds of the global deployed model obtained in {\tt DFL}.
\begin{lemma}[One-step behavior of $e_1^{(t)}$, $e_2^{(t)}$ and $e_3^{(t)}$] \label{lem:An_oneSTP_m}
For $t\in\mathcal T_k$ before performing global synchronization, under Assumptions \ref{beta},~\ref{assump:SGD_noise} and~\ref{assump:sub_err}, if $\eta_{k}\leq\frac{2}{\mu+\beta},~\forall k$, using {\tt DFL} for ML model training, in $t\in\mathcal T_k$, the one-step behaviors of $(e_1^{(t+1)})^2$, $e_2^{(t+1)}$ and $e_3^{(t+1)}$ are presented as follows:
\begin{align} \label{eq:e1_oneSTP}
    &(e_1^{(t+1)})^2
    \leq
    (1-\mu\eta_{k})^2(e_1^{(t)})^2
    +\eta_{k}^2(\sigma^2+\phi^2),
    \\\label{eq:e2_oneSTP}
    &e_2^{(t+1)}\leq
    (1+\eta_{k}(\beta-\mu))e_2^{(t)} 
    +2\omega\eta_{k}\beta e_3^{(t)}
    +\eta_{k}\delta,
\\ \label{eq:e3_oneSTP}
&   e_3^{(t+1)} \leq
     (1-\eta_{k}\mu)e_3^{(t)} 
    +\eta_{k}\beta e_2^{(t)}.
\end{align} 
\end{lemma}
Lemma~\ref{lem:An_oneSTP_m} characterizes the one-step dynamics of $e_1^{(t)}$, $e_2^{(t)}$, and $e_3^{(t)}$ within a local model training interval. During local model updates, the upper bounds of $e_1^{(t)}$ and $e_3^{(t)}$ display a contraction behavior among different terms, while the upper bound of $e_2^{(t)}$ exhibits a monotonic increase. Given that the upper bounds for $e_2^{(t)}$ and $e_3^{(t)}$ are interrelated, it is essential to investigate their mutual impact and the role of the combiner weight in shaping their behavior at the global synchronization stage. This investigation will be conducted in the subsequent proposition.
\begin{proposition}[Upper bounds for $e_1^{(t_{k+1})}$, $e_2^{(t_{k+1})}$ and $e_3^{(t_{k+1})}$] \label{prop:An_oneSTP_t}
    Under Assumptions \ref{beta},~\ref{assump:SGD_noise} and~\ref{assump:sub_err}, if $\eta_{k}=\frac{\eta_{\mathrm{max}}}{1+\gamma k}$, where $\eta_{\mathrm{max}}<\min\left\{\frac{2}{\beta+\mu},\frac{(\tau-\Delta)\mu}{\beta^2[(1+\lambda_+)^{\tau}-1-\tau\lambda_+]}\right\}$, there exists constants $C_1$, $C_2$, $C_3$, $K_1$, $K_2$ and $\lambda_\pm$ such that $(e_1^{(t_{k+1})})^2$, $e_2^{(t_{k+1})}$ and $e_3^{(t_{k+1})}$ across global synchronizations in {\tt DFL} can be bounded as 
\begin{align} \label{eq:e1_main_m}
    (e_1^{(t_{k+1})})^2
    \leq&
    \left(1-\eta_k/\eta_{\mathrm{max}}C_1\right)(e_1^{(t_k)})^2
    +\eta_{k}^2 (\tau-(1-\alpha)\Delta)(\sigma^2+\phi^2),
\end{align} 
\begin{align} \label{eq:e2_main_m}
    e_2^{(t_{k+1})}&\leq 
    \alpha(1+\lambda_+)^{\tau}e_2^{(t_k)}
    +\eta_k\alpha 2\omega C_2e_3^{(t_k)}
    +\eta_k\alpha K_1\delta,
\end{align}
\begin{align} \label{eq:e3_sync_m}
     &  e_3^{(t_{k+1})}\leq
    (1-\eta_k\beta C_3) e_3^{(t_k)}
    +C_2 \eta_ke_2^{(t_k)}
    +\eta_k^2 K_2\delta,
\end{align}
 where the expressions of the constants are provided in Appendix~\ref{app:main_gap}.
\end{proposition}
\begin{skproof}
The complete proof is provided in Appendix~\ref{app:main_gap}. To prove Proposition~\ref{prop:An_oneSTP_t}, we first use Lemma~\ref{lem:An_oneSTP} to derive the one-step dynamics of $e_1^{(t+1)}$, $e_2^{(t+1)}$, and $e_3^{(t+1)}$ across the local model training period. Next, we apply the one-step dynamics from Lemma~\ref{lem:An_oneSTP} repeatedly to solve the coupled dynamics and obtain the recurrence relationship of $e_1^{(t_{k+1})}$, $e_2^{(t_{k+1})}$, and $e_3^{(t_{k+1})}$ across the global synchronization periods as follows:
\begin{align} \label{eq:e1_sync_m}
\hspace{-0.1in}
    (e_1^{t_{k+1}})^2
    \leq [(1-\alpha)(1-\mu\eta_{k})^{2(\tau-\Delta)}+\alpha(1-\mu\eta_{k})^{2\tau}](e_1^{(t_k)})^2
    +[\tau-(1-\alpha)\Delta]\eta_{k}^2(\sigma^2+\phi^2),
    \hspace{-0.1in}
\end{align} 
\begin{align} \label{eq:e2_sync_m}
    e_2^{(t_{k+1})}&\leq 
    \alpha\Pi_{+,t_{k+1}}e_2^{(t_k)}
    +\alpha\frac{4\omega}{\sqrt{8\omega+1}}[\Pi_{+,t_{k+1}}-1]e_3^{(t_k)}
    +\alpha\frac{\mu}{-\beta^2\lambda_+\lambda_-}[\Pi_{+,t_{k+1}}-1]\delta,
\end{align}
\begin{align} \label{eq:e3_init_m}
         &e_3^{(t_{k+1})}\leq
        \Psi_1(\eta_k) e_3^{(t_k)}
        +\underbrace{2g_{3}[(1-\alpha)\Pi_{+,t_{k+1}-\Delta}+\alpha\Pi_{+,t_{k+1}}-1]}_{(a)}e_2^{(t_k)}
        \nonumber\\&
        +\underbrace{\left[(1-\alpha)[g_{5}(\Pi_{+,t_{k+1}-\Delta}-1)+g_{6}(\Pi_{-,t_{k+1}-\Delta}-1)]
        +\alpha[g_{5}(\Pi_{+,t_{k+1}}-1)+g_{6}(\Pi_{-,t_{k+1}}-1)]\right]}_{(b)}\delta/\beta
    \end{align}
with the expression of $\Psi_1(\cdot)$, $g_3$, $g_5$, $g_6$, and $\Pi_{\{+,-\},t}=[1+\eta_{k}\beta\lambda_{\{+,-\}}]^{t-t_{k}}$ provided in Lemma~\ref{lem:main} in Appendix~\ref{app:lemmas}. Utilizing the convexity of $[(1-\alpha)(1-\mu\eta_{k})^{2(\tau-\Delta)}+\alpha(1-\mu\eta_{k})^{2\tau}]$ in~\eqref{eq:e1_sync_m} and $\Pi_{+,t_{k+1}}$~\eqref{eq:e2_sync_m}, we are able to derive the bounds presented in~\eqref{eq:e1_main_m} and~\eqref{eq:e2_main_m}.
 Finally, considering~\eqref{eq:e3_init_m},
    we bound $\Psi_1(\eta_k)$ as follows: 
    Since $\lambda_-\leq\lambda_+$, $\eta_k\leq\eta_{\mathrm{max}}$
    and $\eta_k\beta\leq1$, we apply the binomial theorem along with a set of algebraic manipulations to get
    \begin{align}
        &\frac{\Psi_1(\eta_k)-1}{\eta_k\beta}
        \leq 
        -(\tau-\Delta)\mu/\beta 
        + \eta_{\mathrm{max}}\beta[(1+\lambda_+)^{\tau}-1-\tau\lambda_+]
        \triangleq -C_3,
    \end{align}
    and therefore $\Psi_1(\eta_k)\leq 1-\eta_k\beta C_3<1$. We then proceed by bounding $(a)$ in~\eqref{eq:e3_init_m}. Using the convexity of $\Pi_{+,t}-1$ with respect to both $\eta_k\beta$ and $\eta_{\mathrm{max}}\beta$ with the constraint $\eta_{\mathrm{max}}\beta\leq1$, we obtain
$
        2g_3[(1-\alpha)\Pi_{+,t_{k+1}-\Delta}+\alpha\Pi_{+,t_{k+1}}-1]
        \leq 
        \eta_k C_2.
$
Finally, we bound $(b)$ in~\eqref{eq:e3_init_m}, using the binomial expansion and the expressions of $g_5$ and $g_6$, which yields
    $
    g_{5}(\Pi_{+,t}-1)+g_{6}(\Pi_{-,t}-1)
    \leq \eta_k^2\beta K_2.
    $
Replacing these bounds in~\eqref{eq:e3_init_m} leads to the final result in~\eqref{eq:e3_sync_m}.
\end{skproof}

Proposition~\ref{prop:An_oneSTP_t} provides insight about the impact of subnets on the convergence of model training by bounding the subnet deviation error $(e_1^{(t_{k+1})})^2$ in~\eqref{eq:e1_main_m}. It shows that $(e_1^{(t_{k+1})})^2$ is (i) dependent on the optimality gap of the subnet error-free variable (i.e., $\mathbb E[\Vert\bar{\mathbf v}_c^{(t)}-\mathbf w^*\Vert^2],~\forall t\in\mathcal T_k$ encapsulated in $\phi$), (ii) sensitive to SGD noise and intra-subnet data diversity encapsulated in $\phi$ (the bound increases as $\sigma$, $\delta_c$ and $\omega_c$ increase), and (iii) getting larger as the period of local aggregation (i.e., $t^{\mathsf{L}}_{k}$) increases. To demonstrate the convergence of $(e_1^{(t_{k+1})})^2$, it is necessary to ensure Assumption~\ref{assump:sub_err} is satisfied by monitoring the dynamic of $\mathbb E[\Vert\bar{\mathbf v}_c^{(t)}-\mathbf w^*\Vert^2]$ and selecting appropriate local aggregation instances $\{\Theta_c^{(t)}\}_{t\in\mathcal T_k}$. Our control algorithm in Sec.~\ref{Sec:controlAlg} demonstrates how this can be achieved. Proposition~\ref{prop:An_oneSTP_t} also reveals the evolution of $e_2^{(t_k)}$ and $e_3^{(t_k)}$ across global synchronizations. The bounds~\eqref{eq:e2_main_m} and~\eqref{eq:e3_sync_m} demonstrate that $e_2^{(t_k)}$ and $e_3^{(t_k)}$ are (i) interdependent, forming a coupled relationship, and (ii) influenced by the inter-subnet data diversity (the bounds increase as $\delta$ increases for a fixed value of $\omega$). The bound in \eqref{eq:e2_main_m} also establishes a necessary condition for the combiner weight $\alpha$ to exhibit a contraction behavior during training. In particular, to achieve this behavior, $\alpha$ must be strictly less than $\frac{1}{(1+\lambda_+)^\tau}$. This condition is crucial for {\tt DFL} to achieve convergence and will be further used in Theorem~\ref{thm:subLin_m}.

\subsection{General Convergence Behavior of {\tt DFL}}
\label{ssec:convAvg}

Using the aforementioned results, 
we will next demonstrate that the global model deployed under {\tt DFL} can achieve sublinear convergence to the optimum model.


\begin{theorem}[Sublinear Convergence of {\tt DFL}] \label{thm:subLin_m}
 Under Assumptions \ref{beta},~\ref{assump:SGD_noise} and~\ref{assump:sub_err}, if $\eta_{k}=\frac{\eta_{\mathrm{max}}}{1+\gamma k},~\forall k$ and $\vert\mathcal T_k\vert\leq\tau,~\forall k$, there exists constants $Y_1$, $Y_2$ and $Y_3$ such that the distance between the global model and the optimal model at each global synchronization in {\tt DFL} can be bounded as 
    \begin{align}\label{eq:main_m}
        \mathbb E[\Vert\bar{\mathbf w}^{(t_k)}-\mathbf w^*\Vert^2]
        \leq \underbrace{2Y_1^2 \eta_k}_{(a)}+\underbrace{2Y_3^2\eta_k^2}_{(b)},
    \end{align}
    where 
    $\eta_{\mathrm{max}}<\min\left\{\frac{2}{\beta+\mu},\frac{(\tau-\Delta)\mu}{\beta^2[(1+\lambda_+)^{\tau}-1-\tau\lambda_+]}\right\}$, $\gamma<\min\left\{1-(1-\mu\eta_{\mathrm{max}})^{2(\tau-\Delta)},C_3\eta_{\mathrm{max}}\beta\right\}$,
    \begin{align}
        \alpha<\alpha^* \triangleq \frac{1}{\frac{C_2\eta_{\max}^2}{\eta_{\max}\beta C_3-\gamma}
         2\omega C_2(1+\gamma)
        +(1+\gamma)(1+\lambda_+)^{\tau}},
    \end{align}
    with the expression of the constants provided in Appendix~\ref{app:subLin}.
\end{theorem}

\begin{skproof}
The complete proof is provided in Appendix~\ref{app:subLin}. 
We first obtain that \\
$
    \sqrt{\mathbb E[\Vert\bar{\mathbf w}^{(t_{k})}-\mathbf w^*\Vert^2]} 
    \leq e_1^{(t_k)}+e_3^{(t_k)},
$
and thus
\begin{align}\label{eq:final_m}
    \mathbb E[\Vert\bar{\mathbf w}^{(t_k)}-\mathbf w^*\Vert^2]\leq (e_1^{(t_k)}+e_3^{(t_k)})^2\leq 2(e_1^{(t_k)})^2+2(e_3^{(t_k)})^2.
\end{align}
We will prove~\eqref{eq:main_m} by induction, showing that $e_1^{(t_k)}\leq Y_1\sqrt{\eta_k}$, $e_2^{(t_k)}\leq Y_2\eta_k$, and $e_3^{(t_k)}\leq Y_3\eta_k$. The base of induction trivially holds since $Y_1\geq0$, $Y_2\geq0$, and $Y_3\geq\eta_{\mathrm{max}}e_3^{(0)}$ at the start of training ($k=0$). For the induction step, assume that the statement holds true for some $k \in \mathbb{N}$. We then show that it also holds for $k+1$. 
To show $e_1^{(t_{k+1})}\leq\sqrt{\eta_{k+1}}Y_1$, we use~\eqref{eq:e1_main_m}  and the induction hypothesis ($e_1^{(t_{k})}\leq\sqrt{\eta_{k}}Y_1$), yielding the sufficient condition for all $k\geq 0$:
$$
\eta_{\mathrm{max}}(\tau-(1-\alpha)\Delta)(\sigma^2+\phi^2)
\leq 
[C_1-\gamma]Y_1^2,
$$
which is verified since 
$\gamma<1-(1-\mu\eta_{\mathrm{max}})^{2(\tau-\Delta)}\leq C_1$
 and $Y_1^2=\frac{(\tau-(1-\alpha)\Delta)(\sigma^2+\phi^2)\eta_{\mathrm{max}}}{C_1-\gamma}.$
Thus we can show for $e_1^{(t_k)}$ that $(e_1^{(t_k)})^2\leq\eta_{k}Y_1^2,~\forall k$.
To show $e_2^{(t_{k+1})}\leq\eta_{k+1}Y_2$, we use~\eqref{eq:e2_main_m} and the induction hypothesis ($e_2^{(t_{k})}\leq\eta_{k}Y_2$), yielding 
 \begin{align} 
&\alpha(1+\lambda_+)^{\tau}\eta_k Y_2 
    +\alpha 2\omega C_2\eta_k^2 Y_3
    +\alpha K_1\eta_k\delta
    -\eta_{k+1}Y_2\leq 0,
\end{align} 
To satisfy the above condition for all $k\geq 0$, it is enough to have
 \begin{align}\label{eq:e2_cond2_m}
    & \frac{[1-\alpha(1+\gamma)(1+\lambda_+)^{\tau}]Y_2-\alpha K_1\delta(1+\gamma)}{\eta_{\mathrm{max}}}
    -\alpha 2\omega C_2Y_3(1+\gamma)
    \geq 0.
 \end{align} Holding on proving the final results of the upper bound on $e_2^{(t_{k+1})}$, we take a look at $e_3^{(t_{k+1})}$
 to show $e_3^{(t_{k+1})}\leq\eta_{k+1}Y_3$. We use~\eqref{eq:e3_sync_m} and the induction hypothesis ($e_3^{(t_{k})}\leq\eta_{k}Y_3$), yielding 
\begin{align}
    (1-\eta_k\beta C_3)Y_3\eta_k
    +[C_2Y_2+K_2\delta] \eta_k^2
    - Y_3\eta_{k+1}\leq 0.
\end{align}
The above condition is equivalent to satisfying the following condition for all $k\geq 0$:
\begin{align} \label{eq:cond_Y3_m}
    Y_3[\gamma -\eta_{\max}\beta C_3]
    +[C_2Y_2+K_2\delta]\eta_{\max}
    \leq 0.
\end{align}
To demonstrate $e_2^{(t_k)}\leq\eta_{k+1}Y_2$ and $e_3^{(t_k)}\leq\eta_{k+1}Y_3$, it is necessary that conditions $e_3^{(t_k)}\geq \eta_{\mathrm{max}}e_3^{(0)}$, \eqref{eq:e2_cond2_m}, and~\eqref{eq:cond_Y3_m} hold simultaneously, which can be achieved by noting that $\alpha<\alpha^*$ and utilizing the definition of $Y_2$ in~\eqref{eq:Y2}.
Completing the induction, we show that $e_2^{(t_k)}\leq\eta_{k+1}Y_2$ and $e_3^{(t_k)}\leq\eta_{k+1}Y_3$. Finally, substituting these results back into~\eqref{eq:final_m} completes the proof.
\end{skproof}

Investigating the bound in~\eqref{eq:main_m} of Theorem~\ref{thm:subLin_m}, we found conditions for {\tt DFL} to achieve a convergence rate of $\mathcal{O}(1/t)$. Note that the decay rate of $(a)$ is faster than $(b)$ across global synchronizations, indicating that the impact of inter-gradient diversity (i.e., $\omega$ and $\delta$) incorporated into $Y_3$ on the bound decays faster than that of intra-gradient diversity incorporated into $Y_1$.
The bound also demonstrates the effect of the combiner weight $\alpha$ on convergence. In particular, when $\Delta=0$, the optimal choice of $\alpha$ is the trivial solution (i.e., $\alpha=0$ since this choice leads to minimizing $Y_3$), indicating that it is optimal for {\tt DFL} to perform the standard {\tt FedAvg} algorithm without local-global model combination. Given a fixed value of $\omega$, as the gradient diversity (i.e., $\delta$) increases, it becomes more favorable to choose a smaller value of $\alpha$ to achieve better convergence, demonstrating the importance of putting a higher importance on the global model during local-global model combination to avoid having biased local models under data heterogeneity.  The value of the bound increases with respect to the subnet deviation noise (i.e., $\phi$), implying that {\tt DFL} can achieve the same performance under less frequent global aggregations with more frequent local aggregations. This further suggests that the local model training interval can be prolonged upon performing more rounds of local aggregations in between global aggregations. Moreover, the convergence bound shows the non-triviality of selecting $\alpha$ for the best convergence behavior (as discussed in Sec.~\ref{Sec:controlAlg}).

In the next section, we will leverage these relationships in developing an adaptive control algorithm for {\tt DFL} that tunes the algorithm parameters to achieve the convergence bound in Theorem~\ref{thm:subLin} while reducing the network costs.

\section{Adaptive Control Algorithm for {\tt DFL}}\label{Sec:controlAlg}
\noindent In this section, we develop a control algorithm based on Theorem~\ref{thm:subLin_m} for tuning the controllable parameters in {\tt DFL}, while guaranteeing the sublinear convergence of the model. In {\tt DFL}, there are four sets of controllable parameters: (i) local model training intervals $\{\tau_k\}$, (ii) the combiner weight $\{\alpha_k\}$, (iii) the gradient descent step size $\{\eta_k\}$, and (iv) the instances of local aggregations $\{\mathcal T^\mathsf{L}_{k,c}\}$. The decisions on (i), (ii), (iii) and (iv) are made by the main server at $t=t_{k+1}-\Delta_k^{\mathsf{D}},~\forall k$ during global aggregation. 

To tune these parameters, we develop a control algorithm consisting of the following two parts. \textit{Part I:}  an adaptive technique (Sec.~\ref{subsec:learniParam}) to determine the step-size (i.e., $\eta_{max}$ and $\gamma$ in $\eta_k$ defined in Theorem~\ref{thm:subLin_m}) considering the conditions imposed by Theorem~\ref{thm:subLin_m}. \textit{Part II:} an optimization scheme (Sec.~\ref{subsec:learnduration}) to tune $\tau_k$ and $\alpha_k$ accounting for the tradeoff between the ML model performance and network resource consumption. 
In Sec.~\ref{ssec:control}, we provide the pseudocode summarizing how Parts I and II are integrated.

\subsection{Step Size Parameters ($\eta_{max}$, $\gamma$)}\label{subsec:learniParam}
We first tune the step size parameters ($\eta_{max}$, $\gamma$). This is done for a given set of model-related parameters ($\beta, \mu, \zeta, \delta, \zeta_c, \delta_c, \sigma$, $\omega$ and $\omega_c$), which can be estimated by the server (e.g., see Sec. IV-C of~\cite{lin2021timescale}). 
Given the fact that larger feasible values of $\eta_{max}$ result in larger values of step size and thus faster convergence, given the conditions mentioned in the statement of Theorem~\ref{thm:subLin_m}, we first determine the largest value for $\eta_{max}$ such that $\eta_{max}<\min\left\{\frac{2}{\beta+\mu},\frac{(\tau-\Delta)\mu}{\beta^2[(1+\lambda_+)^{\tau}-1-\tau\lambda_+]}\right\}$, where $\lambda_+$ is defined in Theorem~\ref{thm:subLin_m}. Afterward, 
we arbitrarily choose the value of $\gamma$ such that $\gamma<\min\left\{1-(1-\mu\eta_{\mathrm{max}})^{2(\tau-\Delta)},C_3\eta_{\mathrm{max}}\beta\right\}$.


We next introduce the optimization formulation to determine the length of local model training interval $\tau_k$ and the combiner weight $\alpha_k$ for each local model training interval.

\subsection{Length of Local Training Interval $\{\tau_k\}$ and Value of Combiner Weight $\{\alpha_k\}$}\label{subsec:learnduration}

Considering the convergence goal of {\tt DFL} (i.e., sublinear convergence with low resource consumption across edge devices), we formulate an optimization problem $\bm{\mathcal{P}}$ solved by the main server at each instance of global aggregation at $t=t_{k}-\Delta_{k-1}^{\mathsf{D}},~\forall k$ to tune $\tau_k$ and $\alpha_k$ for the subsequent local model training intervals $\mathcal{T}_k,~\forall k$. The objective function of $\bm{\mathcal{P}}$ accounts for the joint impact of three metrics: $(a)$ energy consumption of local and global model aggregation, $(b)$ communication delay of local and global aggregation, and $(c)$ the performance of global deployed model captured by the optimality gap in Theorem~\ref{thm:subLin_m}. 

\begin{align*} 
    &(\bm{\mathcal{P}}): ~~\min_{\tau,\alpha}  c_1\underbrace{(\frac{T-t_{k}}{\tau})\Big(E_{\textrm{GlobAgg}}+\sum\limits_{c=1}^N \vert\mathcal{A}_{\{c,k\}}\vert E_{c,\textrm{LocalAgg}}\Big)}_{(a)}+
    \nonumber \\&
    c_2\underbrace{(\frac{T-t_{k}}{\tau})\Big(\Delta_{\textrm{GlobAgg}}+\sum\limits_{c=1}^N\vert\mathcal{A}_{\{c,k\}}\vert\Delta_{c,\textrm{LocAgg}}\Big)}_{(b)}
    + c_3\underbrace{\nu(\tau,\alpha)}_{(c)}
\end{align*}  
\vspace{-0.3in}  
    \begin{align} 
    & \textrm{s.t.}\nonumber \\ 
   & \;\;\; \Delta \leq \tau \leq \min{\{\tau_\textrm{max}, T-t_{k}\}}, \tau_k\in\mathbb{Z}^+, \label{eq:tauMax}\\   
   & \;\;\; \alpha<\frac{1}{\frac{C_2\eta_{\max}^2}{\eta_{\max}\beta C_3-\gamma}
         2\omega C_2(1+\gamma)
        +(1+\gamma)(1+\lambda_+)^{\tau}}, \label{eq:bound_alpha}
\end{align}
where $E_{\textrm{c,LocalAgg}}= \sum_{j\in \mathcal{S}_c} M\times Q\times p_{j}/R_j^{(t)}$ is the energy consumption of conducting local model aggregation at edge server $n_c$, where $M$ denotes the size of the model (i.e., number of model parameters), $Q$ denotes the number of bits used to represent each model parameter, which is dependent on the quantization level, $p_{j}$ denotes the transmit power of device $j\in\mathcal S_c$,  and $R_j^{(t)}=W\log_2 \left(1+ \frac{p_j \vert {h}^{(t)}_{j}\vert^2}{N_0W} \right)$ is the transmission rate between device $j\in\mathcal S_c$ and its associated edge server $n_c$ at time $t$. The noise power is $N_0 W$, with $N_0$ as the white noise power spectral density, $W$ as the bandwidth, and ${h}^{(t)}_{j}$ as the channel coefficient.
$E_{\textrm{GlobAgg}}=\sum_{n_c \in\mathcal{N}} M\times Q \times \bar{p}_{n_c}/\bar{R}_{n_c}^{(t)}$ is the energy consumption for edge-to-main server communications, where $\bar{p}_{n_c}$ and $\bar{R}_{n_c}^{(t)}$ denote the transmit power of edge server $n_c$ and the transmission rate between the edge server $n_c\in\mathcal N,~\forall c$ and the main server, respectively. Furthermore, $\Delta_{\textrm{c,LocalAgg}}=\max_{j\in\mathcal{S}_c} \{ M\times Q/R_{j}^{(t)}\}$ is the communication delay of performing local aggregation via device $j\in\mathcal S_c$.\footnote{The device-to-edge server communications are assumed to occur in parallel, using multiple access techniques such as FDMA.}
$\Delta_{\textrm{GlobAgg}}=\max_{i\in\mathcal I}\{\Delta_k / R(\xi_i^{(t)})\}$ is the device-to-main server communication delay, where $\Delta_k$ denotes the round-trip delay measured in terms of the number of conducted SGDs and $R(\xi_i^{(t)})$ denotes the processing rate (the number of SGDs conducted at each time instance measured in seconds) at edge device $i$. 

In $\bm{\mathcal{P}}$, $\vert\mathcal{A}_{\{c,k\}}\vert$ is the number of local aggregations performed by devices $i\in\mathcal S_c$ within a period of local model training interval, which we obtain as $\mathcal{A}_{\{c,k\}} = \{t\in\mathcal T_k:  \sum\limits_{c=1}^N\varrho_c(1-\Theta_c^{(t)})(2\delta_c^2+4\omega_c^2\beta^2\Vert\bar{\mathbf v}_c^{(t)}-\mathbf w^*\Vert^2)>\phi^2 \}$ (see Assumption~\ref{assump:sub_err}). To obtain $\mathcal{A}_{\{c,k\}}$ we thus need to control $\Vert\bar{\mathbf v}_c^{(t)}-\mathbf w^*\Vert^2$, to monitor the value of which, we first approximate it as $\Vert\bar{\mathbf v}_c^{(t)}-\mathbf w^*\Vert^2\approx\Vert\bar{\mathbf w}_c^{(t)}-\mathbf w^*\Vert^2$. Then, during each local aggregation, the edge server estimates the upper bound on $\Vert\bar{\mathbf w}_c^{(t)}-\mathbf w^*\Vert^2$ using strong convexity of $F(\cdot)$ (i.e., $\Vert\nabla F(\bar{\mathbf w}_c^{(t)})\Vert^2 \geq \mu^2\Vert\bar{\mathbf w}_c^{(t)}-\mathbf w^*\Vert^2$). 
Finally, $\nu(\tau,\alpha)=2Y_1^2 \eta_K+2Y_3^2\eta_K^2$ denotes the optimality gap upper bound derived in Theorem~\ref{thm:subLin_m} at time $t_K=T$. To compute $Y_3$, we first approximate $e_3^{(0)}\approx\mathbb \Vert\bar{\mathbf w}^{(0)}-\mathbf w^*\Vert$, and then estimate its using its upper bound $\left\Vert\nabla F(\bar{\mathbf w}^{(0)})\right\Vert \geq \mu\Vert\bar{\mathbf w}^{(0)}-\mathbf w^*\Vert$.

\textbf{Constraints.}
The constraint in~\eqref{eq:tauMax} guarantees that the value of $\tau_k$ is larger than the edge-to-main server communication delay, matching our assumption in Sec.~\ref{subsec:syst3}.
Constraint~\eqref{eq:bound_alpha} is a condition on $\alpha$ described in Theorem~\ref{thm:subLin_m} and ensures that the value of $\alpha$ lies within a range to guarantee the sublinear convergence of the global deployed model.

\textbf{Solution.} Formulation $\bm{\mathcal{P}}$ is a non-convex  mixed-integer programming problem. Due to the complex nature of the problem, we solve it via exhaustive search, performing line search over the integer values of $\tau$ in the range given in~\eqref{eq:tauMax} and obtain the optimum of $\alpha\in(0,1]$ corresponding to each value of $\tau$. Note that, given a value of $\tau$,  $\bm{\mathcal{P}}$  is still non-convex with respect to  $\alpha$. Therefore, we discretize the search space of $\alpha$ and perform a line search over the discretized search space of $\alpha$ for each $\tau$. Based on this approach, the search space of $\bm{\mathcal{P}}$ remains to be small due to the limited ranges/choices of $\tau$ and $\alpha$ (i.e., the time complexity of performing line search over $\tau$ is $\mathcal O(T-t_{k}-\Delta_{k-1})$ and the time complexity of performing line search over $\alpha$ is $\mathcal O( 1/S_{\alpha})$, where $S_{\alpha}$ is the discretization step used to discretize  $(0,1]$ interval, resulting in a time complexity of $\mathcal O((T-t_{k}-\Delta_{k-1})\times1/S_{\alpha})$). Thus, we are able to  solve the problem via a reasonable precision (e.g., $S_{\alpha}=0.01$) in a short duration of time (e.g., less than $3$ seconds on a laptop with Intel(R) Xeon(R) Gold 6242 CPU @ 2.80GHz).
\subsection{{\tt DFL} Control Algorithm}
\label{ssec:control}
The procedure of the {\tt DFL} control algorithm is outlined in Algorithm~\ref{GT}, which integrates the procedures described in Sec.~\ref{subsec:learniParam} and~\ref{subsec:learnduration}. 

\begin{algorithm}[H]
{\scriptsize 
\SetAlgoLined
\caption{{\tt DFL} with adaptive control parameters.} \label{GT}
\KwIn{Desirable subnet deviation error coefficient $\phi$, length of model training $T$} 
\KwOut{Global model $\bar{\mathbf w}^{(T)}$}
 Initialize $\bar{\mathbf w}^{(0)}$ and broadcast it among the edge devices through the edge server.\\
 Initialize estimates of $\zeta \ll 2\beta,\delta,\sigma$.\\
 Initialize $\eta_{max}<\min\left\{\frac{2}{\beta+\mu},\frac{(\tau-\Delta)\mu}{\beta^2[(1+\lambda_+)^{\tau}-1-\tau\lambda_+]}\right\}$ and $\gamma$ for the step size $\eta_k=\frac{\eta_{max}}{1+\gamma k}$ according to Sec.~\ref{subsec:learniParam}.\\ 
Initialize $t=0, ~k=0,~ t_0=0, ~t_1=\tau_0$, with $\tau_0$ chosen randomly, such that $\tau_0\leq T-t_{k},~\forall k$.\\ 
 \While{$t\leq T$}{
     \While{$t\leq t_{k+1}$}{
        \For( // Operation at the subnets){$c=1:N$}
        {
         Each device $i\in\mathcal{S}_c$ performs a local SGD update based on~\eqref{eq:SGD} and~\eqref{8} using $\mathbf w_i^{(t-1)}$ to obtain~$\widetilde{\mathbf{w}}_i^{(t)}$.\\
        \uIf{$t\in\mathcal T^\mathsf{L}_{k,c}$}{
        Devices inside the subnet conduct local model aggregation using~\eqref{eq:local_agg} to obtain the updated local model $\mathbf w_i^{(t)}$.
        }
        \uElseIf{$t=t_{k+1}$}{
        Devices inside each subnet perform global synchronization using~\eqref{eq:aggr_alpha}.
        }
        \Else{
        Each device $i\in\mathcal{S}_c$ obtains its updated local model as $\mathbf{w}_i^{(t)} = \widetilde{\mathbf{w}}_i^{(t)}$
        }
      }
      \uIf( // Operation at the edge server){$t=t_{k+1}-\Delta_k$}{
      Each edge server $n_c$ sends $\mathbf w_{i}^{(t_{k+1}-\Delta_k)}$, $\widehat{\mathbf g}_{i}^{(t_{k+1}-\Delta_k)},~\forall i\in\mathcal S_c$ to the main server.\\
       }
    \ElseIf( // Operation at the main server){$t=t_{k+1}-\Delta_k^{\mathsf{D}}$}{
      Compute $\bar{\mathbf w}^{(t_{k+1}-\Delta_k)}$ according to~\eqref{15}, $\hat{\beta}_k$, $\hat{\mu}_k$, $\hat{\sigma}_k$.\\ 
      Set $\hat{\zeta}_k\ll 2\hat{\beta}_k$ and $\hat{\zeta}_{c,k}\ll 2\hat{\beta}_k$, then compute $\hat{\delta}_k$ and $\hat{\delta}_{c,k}$ using the method in~\cite{lin2021timescale}. \\ 
      Characterize $\eta_{max}$ and $\gamma$ for the step size $\eta_k=\frac{\eta_{max}}{1+\gamma k}$ according to Sec.~\ref{subsec:learniParam}.\\
     Compute the instances of local aggregation for each cluster $c$ using Sec.~\ref{subsec:learnduration}.\\
     Solve the optimization~$\bm{\mathcal{P}}$ to obtain $\tau_{k+1}$ and $\alpha_{k+1}$.\\
     Broadcast (i) $\bar{\mathbf w}^{(t_{k+1}-\Delta_k)}$, (ii) $\alpha_{k+1}$ and (iii) $\eta_k$ among the devices.
    }
     $t\gets t+1$
 }
 $k\gets k+1$ and $t_{k+1}\gets t_k+\tau_k$\\
}
}
\end{algorithm}

\vspace{-1mm}
\section{Experimental Evaluation}
\label{sec:experiments}

\noindent This section presents numerical experiments to validate the performance of {\tt DFL}.
In Sec.~\ref{ssec:setup}, we provide the simulation setup. Then, in Sec.~\ref{ssec:conv-eval}, we study the model training performance and convergence behavior of {\tt DFL} with set control parameters (i.e., $\tau_k$, $\mathcal{T}^{\mathsf{L}}_{k,c}$, and $\alpha_k$), revealing the importance of addressing the hierarchical FL architecture and tuning the combiner weight in the presence of communication delay. In Sec.~\ref{ssec:control-eval}, we compare the convergence behavior of our {\tt DFL} control algorithm with baslines in FL~\cite{wang2019adaptive,Li}, verify our theoretical results (Sec.~\ref{ssec:convAvg}), and illustrate the improvements in resource efficiency achieved by the {\tt DFL} control algorithm. 
\vspace{-2mm}
\subsection{System Setup}
\label{ssec:setup}

We consider a network of $50$ edge devices distributed across $10$ equally-sized subnets, with $5$ devices per subnet. In Sec.~\ref{ssec:control-eval}, devices within each subnet are uniformly distributed within a $30$ m $\times$ $30$ m square field, and the base station is located at the center. 

\textbf{Communication parameters} For device-to-edge wireless communications, we assume a transmission power of $p_i= 24$ dBm for device $i$, $W=1$ MHz bandwidth, and $N_0=-173$ dBm/Hz white noise spectral density. Fading and pathloss are modeled based on~\cite{tse2005fundamentals}, using $h^{(t)}_{i}= \sqrt{\beta_{i}^{(t)}} u_{i}^{(t)}$, where $\beta_{i}^{(t)} = \beta_0 - 10\widehat{\alpha}\log_{10}(d^{(t)}_{i}/d_0)$ is the large-scale pathloss coefficient and $u_{i}^{(t)} \sim \mathcal{CN}(0,1)$ is Rayleigh fading. Here, $\beta_0=-30$ dB is the pathloss at the reference distance of $d_0=1$ m, $\widehat{\alpha}=3.75$ is the pathloss exponent, and $d_i^{(t)}$ is the device-to-edge server distance determined by their proximity. Channel reciprocity is assumed for simplicity. For edge-to-cloud wired communication, we employ a transmission power of $\bar{p}_{n_c}=38$ dBm, $\bar{R}_{n_c}^{(t)}=100$ Mbps data rate, and a $50$ ms propagation delay~\cite{Lin}.

\textbf{Computation parameters} Each device's local computation time for performing each mini-batch SGD iteration on the ML model at time $t$ can be modeled as~\cite{PSL2022federated} $T^{(t)}_{i,Comp}=\max_{i\in\mathcal I}a_i\vert\xi_i^{(t)}\vert/f_i^{(t)}$, where every device's CPU frequency is $f_i^{(t)}=15.36$ MHz, mini-batch size is $\vert\xi_i^{(t)}\vert=128$ data points, and number of CPU cycles needed to process one datapoint is $a_i=600$. The computation energy consumption of each device is modeled as $E^{(t)}_{i,Comp}=(Z_i/2)a_iB_i^{(t)}(f_i^{(t)})^2$, where the effective chipset capacitance~\cite{Tran2019cap} is $Z_i = 2\times10^{-22}$ for all devices. With these parameters, the resultant local model update processing rates are $R(\xi_i^{(t)})=200$ updates/s for all devices. In Sec.~\ref{ssec:ctrl_DFL}, the round-trip delay $\Delta_k$ (measured in SGD iterations) will vary according to the particular realizations generated in each round $k$.

We employ the Fashion-MNIST (F-MNIST) dataset for image classification, comprising samples from $10$ different labels of fashion products. The dataset includes a total of $70$K images, with $60$K designated for training and $10$K for testing. 
We distribute the dataset across devices in a non-i.i.d. manner such that each device has data points exclusively from $s$ of the $10$ total labels. This results in an inherent data skewness across devices which is commonly employed to simulate non-i.i.d. settings in federated learning~\cite{lin2021timescale,wang2021device,hosseinalipour2020multi}.  By default, we set $s = 3$.
We evaluate {\tt DFL} on two models: support vector machine (SVM) with regularized squared hinge loss 
and a convolutional neural network (CNN) with softmax and cross-entropy loss, 
both having model dimension $M = 7840$. 
This allows us to assess {\tt DFL}'s performance on loss functions that are strongly convex (SVM) and those exhibiting non-convex properties (CNN).

\subsection{{\tt DFL} Model Training Performance and Convergence}
\label{ssec:conv-eval}

We first discuss how the two major design aspects of {\tt DFL} (i.e., hierarchical FL framework (Sec.~\ref{subsubsec:hierarchical}) and local-global combiner (Sec.~\ref{subsubsec:combiner})) can affect the speed of ML model convergence and mitigate the effect of delay. We consider {\tt FedAvg}~\cite{mcmahan2017communication}, which performs global aggregations after each round of training with $\tau=1$, as our benchmark. This represents an upper bound on the learning performance as it mimics centralized model training.
\begin{figure}[t]
\includegraphics[width=1.0\columnwidth]{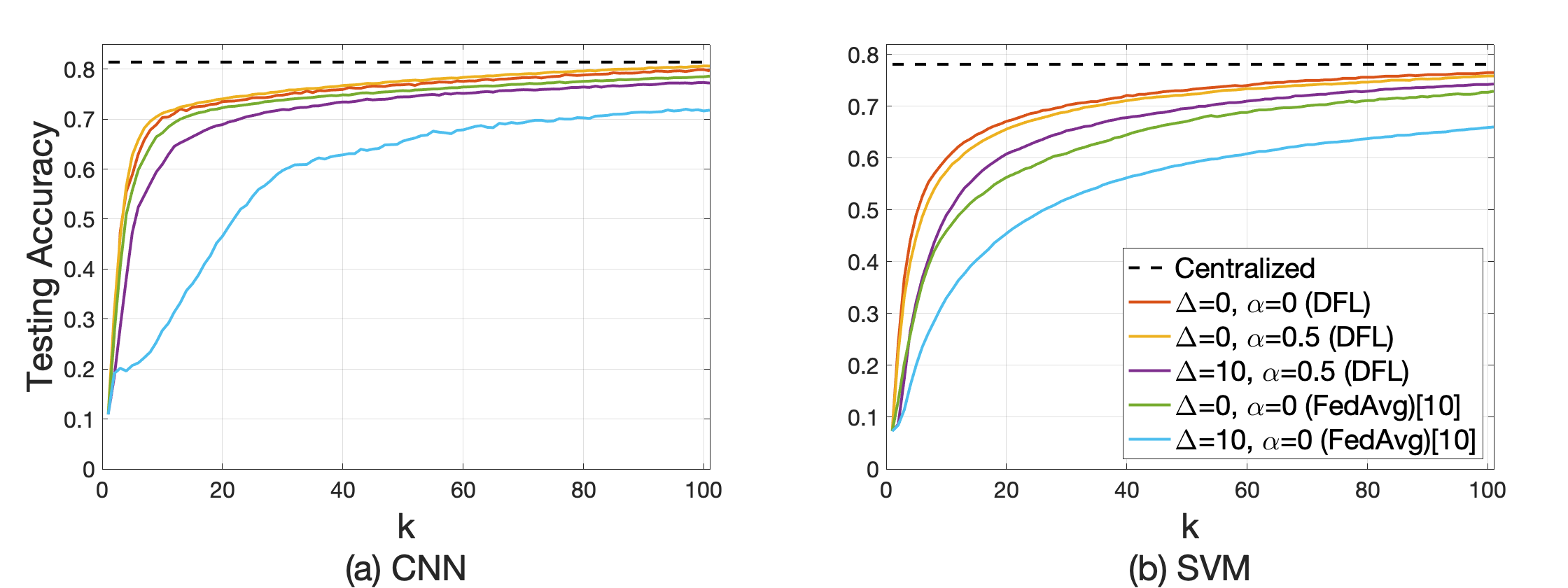}
\centering
\caption{Performance comparison between {\tt DFL} and standard {\tt FedAvg}~\cite{mcmahan2017communication} under various choices of delay: {\tt DFL} surpasses standard {\tt FedAvg} with accuracy gains of $2\%$ (CNN) and $4\%$ (SVM) when delay is negligible ($\Delta=0$), and further outperforms with gains of $6\%$ (CNN) and $8\%$ (SVM) when delay is non-negligible ($\Delta=10$).}
\label{fig:mnist_poc_1_all}
\vspace{-5mm}
\end{figure} 

 
\subsubsection{Model convergence of hierarchical FL} \label{subsubsec:hierarchical}
We compare the performance of {\tt DFL}  (Algorithm~\ref{DFL}) with standard {\tt FedAvg} model aggregation/synchronization~\cite{mcmahan2017communication}. In standard {\tt FedAvg}, we consider a (hypothetical) scenario that the edge devices are directly connected to the main server. Thus the comparison would reveal the impact of hierarchical structure of {\tt DFL} on model training. We consider two different scenarios: (i) when delay is negligible, and (ii) when delay is non-negligible. In scenario (i), we set the combiner weight of {\tt DFL} to $0$ (i.e., $\alpha=0$) with delay $\Delta=0$. In scenario (ii), we set the combiner weight of {\tt DFL} as $\alpha=0.5$ with delay $\Delta=10$. For both scenarios, {\tt DFL} establishes a local model training interval $\tau_k = \tau = 20$ and performs local aggregation in the subnets after each edge device does $5$ local SGD updates.

Fig.~\ref{fig:mnist_poc_1_all} validates the benefit of introducing hierarchical FL with local model aggregations by showing that {\tt DFL} outperforms the standard {\tt FedAvg} under both negligible and non-negligible delay. In scenarios where the delay is insignificant (i.e., $\Delta=0$), {\tt DFL} exhibits superior performance over {\tt FedAvg}, achieving an accuracy gain of $2\%$ for CNN and $4\%$ for SVM. This highlights the benefits of a hierarchical model training structure within an edge-to-cloud network, where frequent local aggregations prevent non-i.i.d. data-driven deviation of local models from the optimum within each subnet. On the other hand, in situations where the delay is non-negligible (i.e., $\Delta=10$), {\tt DFL} outperforms {\tt FedAvg} with accuracy gains of $6\%$ for CNN and $8\%$ for SVM. The result demonstrates the benefits of integrating the local-global combiner with the hierarchical model training architecture to account for the impact of delay.

\begin{figure}[t]
 \includegraphics[width=1.0\columnwidth]{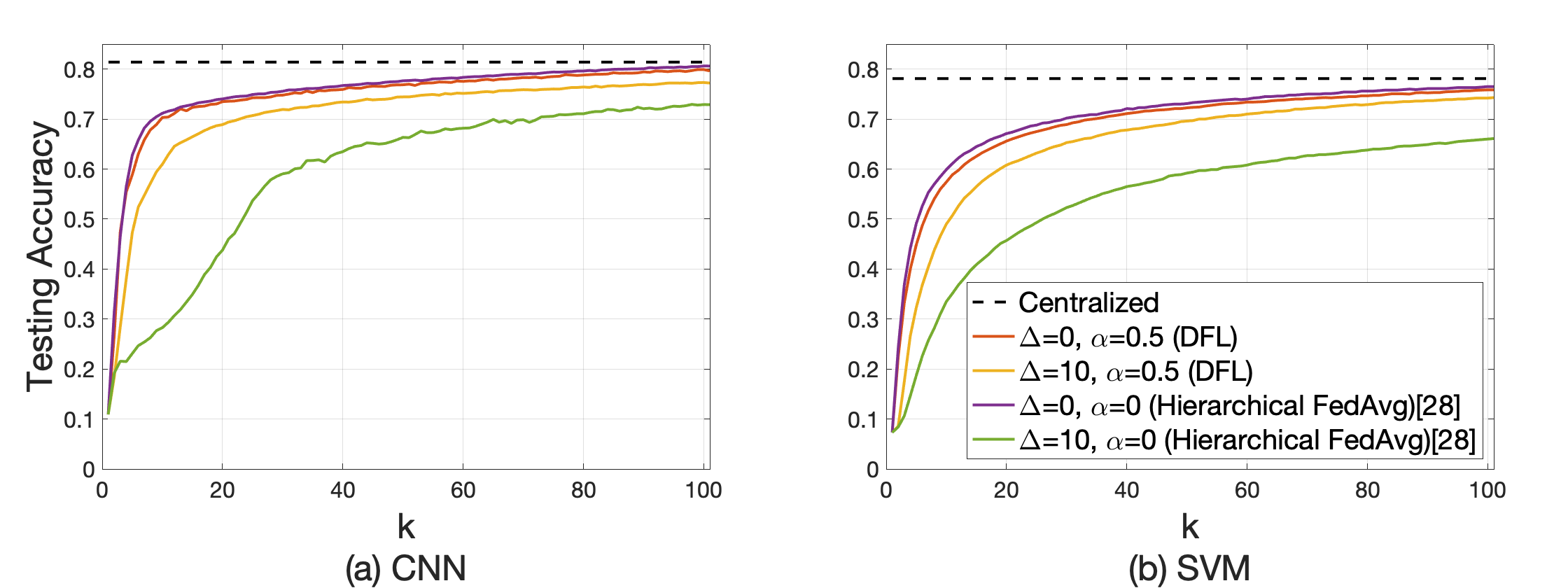}
\centering
\caption{Performance comparison between {\tt DFL} and hierarchical {\tt FedAvg} for various $\alpha$ values in {\tt DFL}: With $\alpha=0.5$, {\tt DFL} significantly outperforms hierarchical {\tt FedAvg} when round-trip delay is $\Delta=10$. Hierarchical {\tt FedAvg} ({\tt DFL} with $\alpha=0$) performs better than {\tt DFL} without delay (i.e., $\Delta=0$), in line with Theorem~\ref{thm:subLin_m}.}
\label{fig:mnist_poc_2_all-2} 
\vspace{-5mm} 
\end{figure}

\subsubsection{Model convergence under local-global combiner} \label{subsubsec:combiner}
To further examine the efficiency enhancements provided by DFL, we conduct a comparative analysis between the performance of {\tt DFL} employing a local-global combiner and the hierarchical {\tt FedAvg} model as outlined in~\cite{liu2020client,Feng2022HFL,Lim2021HFL,Xu2022HFL}, with both methods leveraging local aggregations. Hierarchical {\tt FedAvg} can be thought of as a special case of {\tt DFL} when $\alpha=0$. {\tt DFL} is executed with a fixed combiner weight of  $\alpha=0.5$, which places equal emphasis on the stale global model and up-to-date local models. The local model training interval is set to $\tau_k = \tau = 20$ for both {\tt DFL} and hierarchical {\tt FedAvg}, with a delay of $\Delta=10$. Local aggregations are conducted after each edge device does $5$ local model updates. Furthermore, we plot the convergence behavior of {\tt DFL} when $\alpha=1$, which implies that global model is never used in local devices. 

Fig.~\ref{fig:mnist_poc_2_all-2} shows that {\tt DFL} outperforms vanilla hierarchical {\tt FedAvg} by utilizing the local-global combiner and achieves an accuracy gain of $5\%$ for CNN and $8\%$ for SVM when the delay is large (i.e., $\Delta=10$). Conversely, Fig.~\ref{fig:mnist_poc_2_all-2} illustrates that when there is no delay (i.e., $\Delta=0$), hierarchical {\tt FedAvg} achieves better convergence performance than {\tt DFL}, with an accuracy increase of $1\%$ for both CNN and SVM. These findings align with the result in Theorem~\ref{thm:subLin_m}, suggesting that $\alpha=0$ is the optimal choice of the combiner weight when $\Delta=0$. 
In fact, in {\tt DFL}, $\alpha=0$ simulates the conventional hierarchical {\tt FedAvg}: our results indicate that, when no delay exists ($\Delta=0$), the best approach in {\tt DFL} is to emulate hierarchical {\tt FedAvg}. This validates that the conventional global aggregation procedure is indeed the most effective in the absence of delay. As the delay increases, the choice of $\alpha$ becomes essential and is shaped by several system variables.
\begin{figure}[t]
\includegraphics[width=1.0\columnwidth]{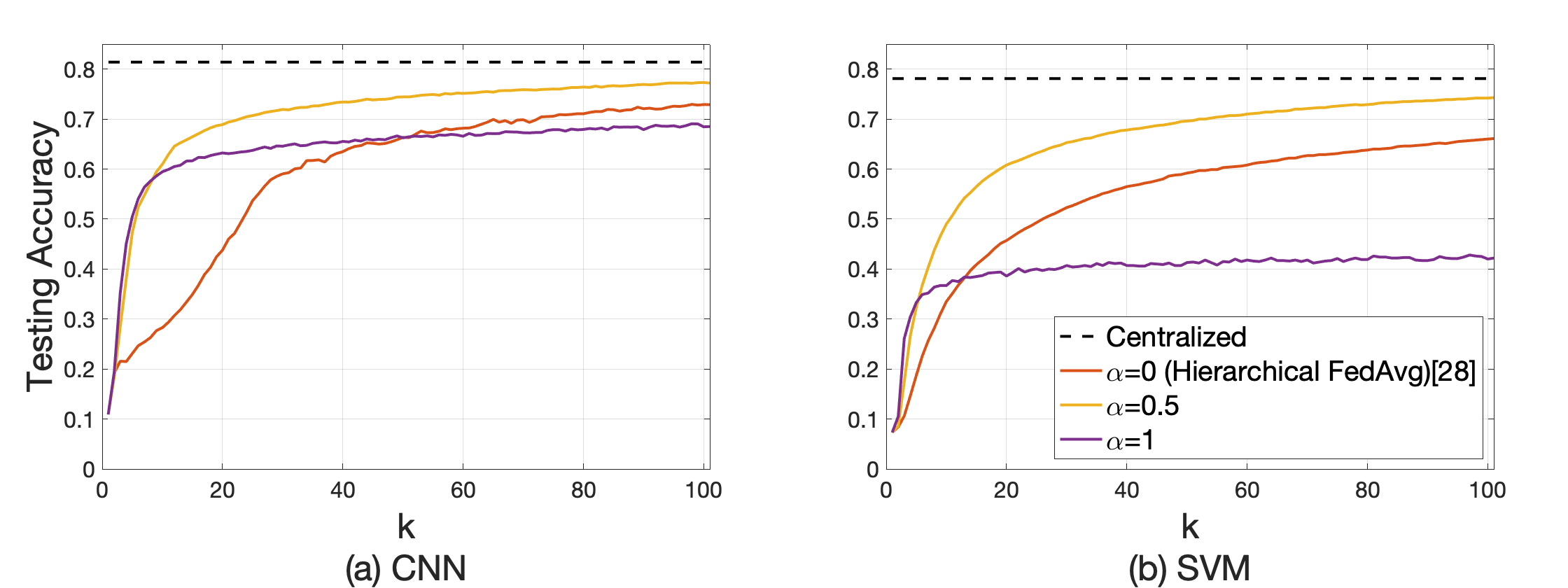} 
\centering
\caption{Performance comparison between {\tt DFL} and hierarchical {\tt FedAvg} under various choice of $\alpha$ in {\tt DFL} under delay $\Delta=10$. With $\alpha=0.5$, {\tt DFL} significantly outperforms hierarchical {\tt FedAvg}. However, the performance of {\tt DFL} was the worst with $\alpha=0$, as no global synchronization was conducted.}
\label{fig:mnist_poc_2_all}
\vspace{-5mm}
\end{figure}
To highlight the robustness of {\tt DFL} against delays, Fig.~\ref{fig:mnist_poc_2_all-2} compares its performance with hierarchical {\tt FedAvg} with a negligible delay (i.e., $\Delta=0$) as the benchmark under a non-negligible delay (i.e., $\Delta=10$). The figure shows that even when the delay is significant, {\tt DFL} achieves an accuracy  within $1\%$ and $2\%$  of the benchmark for CNN and SVM after $100$ global aggregations, demonstrating its delay-robustness.
Lastly, Fig.~\ref{fig:mnist_poc_2_all} shows the convergence performance of {\tt DFL} with no usage of global model (i.e., $\alpha=1$), where the ML model plateaus after reaching a low accuracy. This verifies the condition in Theorem~\ref{thm:subLin_m}, implying that {\tt DFL} may not guarantee sublinear convergence when $\alpha=1$. 

The performance obtained by {\tt DFL} originates from the introduction of the linear local-global combiner during global synchronization. This approach enables the synchronization process to simultaneously consider the outdated yet more generalized global model and the up-to-date yet potentially overfitted local model. Particularly in circumstances where delays are substantial, the system will benefit from preserving a portion of the local model instead of fully synchronizing it with an outdated global model. This approach ensures that the most timely insights derived from the local models are maintained.

\subsection{Adaptive Parameter Control for {\tt DFL}}
\label{ssec:control-eval}
Next, we  analyze the behavior of {\tt DFL} through parameter tuning described in Algorithm~\ref{GT}.
\begin{figure}[t]
\includegraphics[width=1.0\columnwidth]{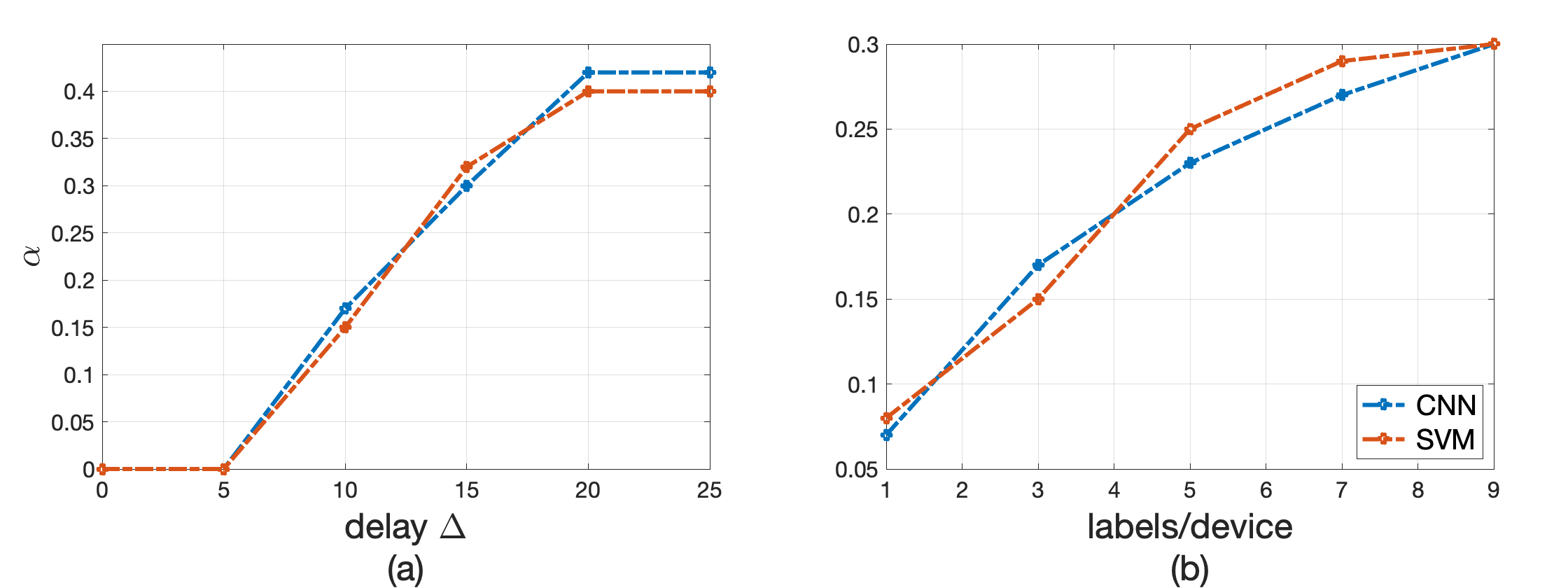}
\centering
\caption{Impact of delay and data diversity on $\alpha$: $\alpha$ increases with delay and decreases with data diversity, indicating that {\tt DFL} emphasizes local models during global synchronization with high delay and global models with high data diversity.}
\label{fig:alpha}
\vspace{-5mm}
\end{figure} 
\subsubsection{Impact of delay on the choice of $\alpha$}
We explore the effect of various values of delay on the selection of the combiner weight $\alpha$. The delay is increased incrementally from $\Delta=5$ to $\Delta=25$ in steps of $5$, while $\tau_k=\tau=30$ is kept constant. Fig.~\ref{fig:alpha}(a) illustrates the average value of $\alpha$ generated by Algorithm~\ref{GT} across global synchronizations. It is evident from the figure that $\alpha$ increases as the delay increases for both CNN and SVM. This aligns with the intuition that as the delay increases, {\tt DFL} would place more emphasis on the local model at the instance of local model synchronization with the global model since the global model becomes more obsolete as the delay increases. 
Furthermore, as delay reaches a threshold, the selection of $\alpha$ ceases to increase due to the constraint in $(\bm{\mathcal{P}})$, limiting it to a feasible range. Algorithm~\ref{GT}'s ability to ensure feasible $\alpha$ values is vital for the convergence behavior of {\tt DFL} described in Theorem~\ref{thm:subLin_m}. 

\begin{figure}[t]
\includegraphics[width=1.\columnwidth]{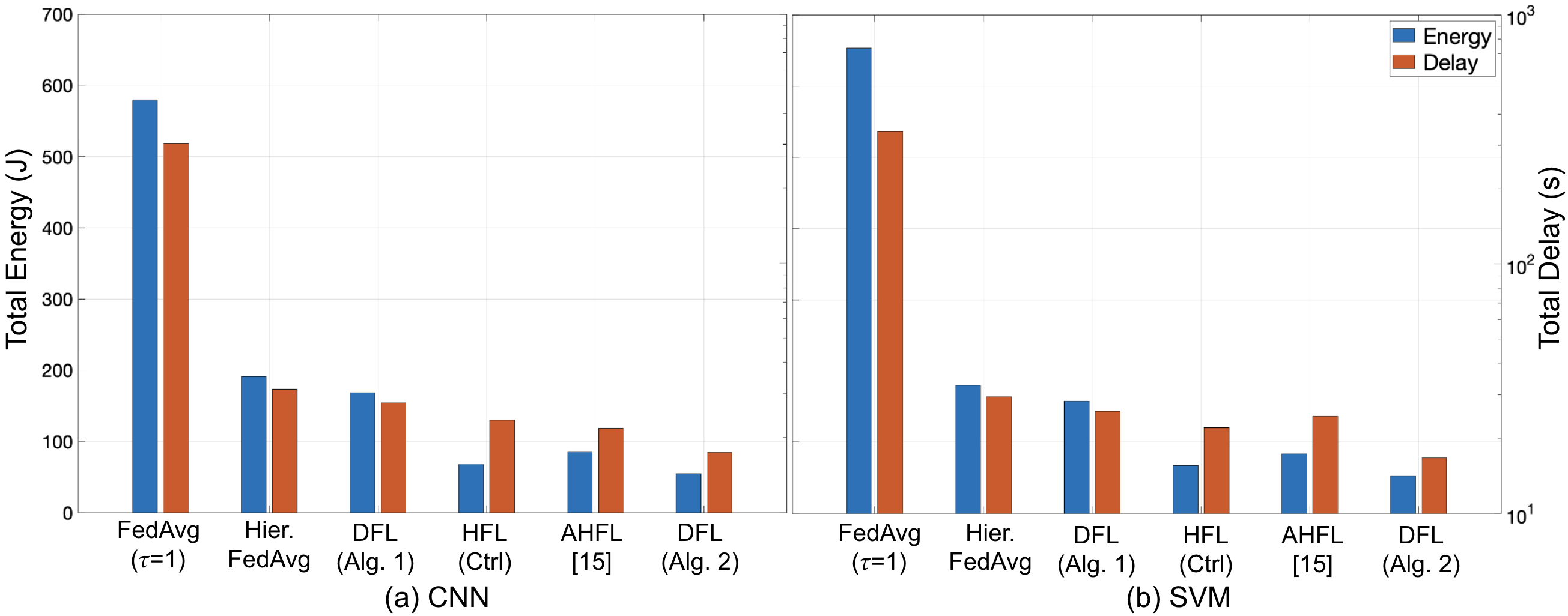}
\centering
\caption{Comparison of {\tt DFL} with adaptive parameter control (Algorithm~\ref{GT}) to the baselines in terms of total energy and delay incurred to reach 80\% testing accuracy. We see that {\tt DFL} obtains substantial improvements in both metrics for both CNN and SVM models.}
\label{fig:res}
\vspace{-5mm}
\end{figure}
\subsubsection{Impact of data diversity on the choice of $\alpha$}
We  explore the impact of levels of data diversity across edge devices on the choice of combiner weight $\alpha$ obtained under Algorithm~\ref{GT}. We increase the data diversity by reducing the number of labels in each device's local dataset. We consider $\Delta=10$ and $\tau_k=\tau=30$. Fig.~\ref{fig:alpha}(b) indicates that $\alpha$ decreases as data diversity increases for both CNN and SVM. This is because
local models diverge more from each other when data diversity is high, making the global model's contribution more critical (i.e., reduction in $\alpha$) in aligning the local models of the devices. 




\subsubsection{Enhanced resource efficiency in comparison to baseline methods}\label{ssec:ctrl_DFL}
The results presented in Fig.~\ref{fig:res} compare the performance of {\tt DFL} with adaptive parameter control (Algorithm~\ref{GT}) with five baseline approaches: (i) FL with full device participation and $\tau=1$; (ii) hierarchical FL with $\alpha=0$, $\tau=20$, where local aggregations are performed after every $5$ local model updates; (iii) {\tt DFL} with fixed parameters (Algorithm~\ref{DFL}) with $\alpha=0.5$, $\tau=20$, where local aggregations are performed after every $5$ local model updates; (iv) {\tt HFL} with parameter control, constructed by setting $\alpha=0$ in the DFL control algorithm, thereby reducing the
DFL’s global aggregation to the standard global aggregation method in~\cite{Feng2022HFL,Lim2021HFL,Xu2022HFL,Luo2020HFEL,wang2021HFL,Mhaisen2022HFL} and (v) Adaptive Hierarchical Federated Learning (AHFL), which incorporates an adaptive control framework to efficiently manage resources~\cite{Xu2022HFL}. Two metrics are used to compare the performance: (M$_1$) total energy consumption and (M$_2$) total delay, each measured upon reaching $80\%$ testing accuracy. For (M$_1$), the results indicated by the blue bars of Fig.~\ref{fig:res}(a) \& (b) show that {\tt DFL} with adaptive parameter control significantly outperforms baseline approaches. Specifically, {\tt DFL} requires $91.8\%$ and $90.5\%$ less energy than baseline (i), $71.4\%$ and $70.4\%$ less energy than baseline (ii), $67.5\%$ and $66.1\%$ less energy than baseline (iii), $19.9\%$ and $21.1\%$ less energy than baseline (iv), and $35.7\%$ and $36.3\%$ less energy than baseline (v) for CNN and SVM models, respectively. Similarly, for (M$_2$), {\tt DFL} requires substantially less communication delay than both baseline approaches as shown in the red bars of Fig.~\ref{fig:res}(a) \& (b). Specifically, {\tt DFL} requires $94.2\%$ and $95.1\%$ less delay than baseline (i), $44.1\%$ and $43.2\%$ less delay than baseline (ii), $36.6\%$ and $35.1\%$ less delay than baseline (iii) $25.8\%$ and $24.5\%$ less delay than baseline (iv), and $20\%$ and $22.5\%$ less delay than baseline (v) for CNN and SVM models, respectively. These results demonstrate the improvements in resource-efficiency provided by {\tt DFL} with adaptive parameter control, attributed to its parameter tuning approach that concurrently considers the tradeoff between the optimality gap, as derived in Theorem~\ref{thm:subLin_m}, the communication delay, and the energy consumption. The improvement of $20-25\%$ over baseline (iv) in both metrics in particular highlights the benefit provided by our local-global model combiner strategy.

\section{Conclusion and Future Work}
\noindent In this work, we proposed {\tt DFL}, which is a novel methodology that aims to improve the efficiency of distributed machine learning model training by mitigating the round-trip communication delay between the edge and the cloud. {\tt DFL} quantifies the effects of delay and modifies the FL algorithm by introducing a linear local-global model combiner used in the local model synchronization steps.
We investigated the convergence behavior of {\tt DFL} under a generalized data heterogeneity metric and obtained a set of conditions to achieve sub-linear convergence. Based on these characteristics, we developed an adaptive control algorithm that adjusts the learning rate, local aggregation rounds, combiner weight, and global synchronization periods. Our numerical evaluation showed that {\tt DFL} leads to a faster global model convergence, lower resource consumption, and a higher robustness against communication delay compared to existing FL algorithms.
Future research directions include improving the robustness of DFL against different types of network impairments, such as jitter and packet loss, and investigating its performance under flexible device participation.

\bibliographystyle{IEEEtran}
\bibliography{ref}

\pagebreak

\begingroup
\let\clearpage\relax 
\onecolumn

\appendices 
\setcounter{lemma}{0}
\setcounter{proposition}{0}
\setcounter{theorem}{0}
\setcounter{definition}{0}
\setcounter{assumption}{0}

\section*{Introduction to Notations and Preliminaries used in the Proofs}\label{app:notations}
\noindent The subnet noise-free variable before global synchronization is introduced as follows:
\begin{equation}\label{eq:v_c}
    \bar{\mathbf v}_c^{(t+1)}= \bar{\mathbf v}_c^{(t)}-\eta_{k}\nabla \bar{F}_c(\bar{\mathbf v}_c^{(t)}),~\forall t\in\mathcal{T}_k\setminus{\{t_k\}},
\end{equation}
with the subnet noise-free variable at global synchronization is defined as 
\begin{align}
    \bar{\mathbf v}_c^{(t_{k+1})}
    =& (1-\alpha)\bar{\mathbf v}^{(t_{k+1}-\Delta)}
    +\alpha\widetilde{\mathbf v}_c^{(t_{k+1})},
\end{align}
where $\widetilde{\mathbf v}_c^{(t_{k+1})}$ is the noise-free variable right before global synchronization, as opposed to $\mathbf v_c^{(t_{k+1})}$ defined right after global synchronization. Similarly, the global noise-free variable is defined as  
\begin{equation}
    \bar{\mathbf v}^{(t+1)}=\sum\limits_{d=1}^N\varrho_{d}\bar{\mathbf v}_d^{(t+1)}~~\forall t\in\mathcal{T}_k.
\end{equation}
The following noise terms used in the appendices are defined as follows:
\begin{align}
 &\label{eq:def_e1}   e_1^{(t)}\triangleq\Big(\mathbb E\Big[\sum\limits_{c=1}^N\varrho_c\sum\limits_{j\in\mathcal S_c}\rho_{j,c}\Vert\mathbf w_i^{(t)} - \bar{\mathbf v}_c^{(t)}\Vert^2\Big]\Big)^{1/2},\\
&\label{eq:def_e2}       e_2^{(t)} \triangleq\sum\limits_{c=1}^N\varrho_{c}\Vert\bar{\mathbf v}_c^{(t)}-\bar{\mathbf v}^{(t)}\Vert,
\\
&\label{eq:def_e3}
        e_3^{(t)}\triangleq\Vert\bar{\mathbf v}^{(t)}-\mathbf w^*\Vert.
    \end{align}
\section{Proof of Theorem~\ref{thm:subLin}} \label{app:subLin} 
\begin{theorem} \label{thm:subLin}
    Under Assumptions \ref{beta},~\ref{assump:SGD_noise} and~\ref{assump:sub_err}, if $\eta_{k}=\frac{\eta_{\mathrm{max}}}{1+\gamma k},~\forall k$ and $\vert\mathcal T_k\vert\leq\tau,~\forall k$, using {\tt DFL} for ML model training, the distance between the global model and the optimum at global synchronization can be bounded as 
    \begin{align}
        \mathbb E[\Vert\bar{\mathbf w}^{(t_k)}-\mathbf w^*\Vert^2]
        \leq 2Y_1^2 \eta_k+2Y_3^2\eta_k^2,
    \end{align}
    where 
    $\eta_{\mathrm{max}}<\min\left\{\frac{2}{\beta+\mu},\frac{(\tau-\Delta)\mu}{\beta^2[(1+\lambda_+)^{\tau}-1-\tau\lambda_+]}\right\}$, $\gamma<\min\left\{1-(1-\mu\eta_{\mathrm{max}})^{2(\tau-\Delta)},C_3\eta_{\mathrm{max}}\beta\right\}$,
    \begin{align}
        \alpha<\alpha^* \triangleq \frac{1}{\frac{C_2\eta_{\max}^2}{\eta_{\max}\beta C_3-\gamma}
         2\omega C_2(1+\gamma)
        +(1+\gamma)(1+\lambda_+)^{\tau}},
    \end{align}
    \begin{align}\label{eq:Y1}
        Y_1 \triangleq\sqrt{\frac{(\tau-(1-\alpha)\Delta)(\sigma^2+\phi^2)\eta_{\mathrm{max}}}{C_1-\gamma}},
    \end{align}
    \begin{align}\label{eq:Y2}
    Y_2\triangleq\max\left\{
    \frac{\eta_{\mathrm{max}}^2\alpha 2\omega C_2(1+\gamma)
    e_3^{(0)}
    +\alpha K_1\delta(1+\gamma)
    }{1-\alpha(1+\gamma)(1+\lambda_+)^{\tau}},
    \frac{\frac{ K_1\delta}{\eta_{\mathrm{max}} 2\omega C_2}
    +\frac{K_2\delta}{\beta C_3-\gamma}
    }{
    \frac{[1-\alpha(1+\gamma)(1+\lambda_+)^{\tau}]}{\eta_{\mathrm{max}}\alpha 2\omega C_2(1+\gamma)}
    -\frac{C_2}{\beta C_3-\gamma}
    }\right\},
\end{align}
    \begin{align}\label{eq:Y3}
        Y_3 \triangleq \max\left\{\eta_{\mathrm{max}}e_3^{(0)},
        \frac{[C_2Y_2+K_2\delta]\eta_{\max}}{\eta_{\max}\beta C_3-\gamma}\right\}.
    \end{align}
    with $K_1=\frac{\mu}{-\beta\lambda_+\lambda_-}[(1+\lambda_+)^{\tau}-1]$, $K_2=\frac{\beta}{\sqrt{1+8\omega}}\sum_{\ell=0}^{\tau-2}\Big(\begin{array}{c}\tau\\\ell+2\end{array}\Big)[\lambda_+^{\ell+1}-\lambda_-^{\ell+1}]$, $C_1=1-((1-\alpha)(1-\mu\eta_{\mathrm{max}})^{2(\tau-\Delta)}+\alpha(1-\mu\eta_{\mathrm{max}})^{2\tau})$, $C_2=\frac{2\beta}{\sqrt{8\omega+1}}[(1+\lambda_+)^{\tau}-1]$, $C_3=(\tau-\Delta)\mu/\beta 
        - \eta_{\mathrm{max}}\beta[(1+\lambda_+)^{\tau}-1-\tau\lambda_+]$ and $\lambda_{\pm} =\frac{1}{2}-\frac{\mu}{\beta}\pm\frac{\sqrt{8\omega+1}}{2}$.
\end{theorem}
\begin{proof}
Note that
\begin{align} 
    &\sqrt{\mathbb E[\Vert\bar{\mathbf w}^{(t_{k})}-\mathbf w^*\Vert^2]} 
    =\sqrt{\mathbb E[\Vert\bar{\mathbf w}^{(t_{k})}-\bar{\mathbf v}^{(t_{k})}+\bar{\mathbf v}^{(t_{k})}-\mathbf w^*\Vert^2]} 
    \\&
    \leq
    \sqrt{\mathbb E[\Vert\bar{\mathbf w}^{(t_{k})}-\bar{\mathbf v}^{(t_{k})}\Vert]}+\Vert\bar{\mathbf v}^{(t_{k})}-\mathbf w^*\Vert
    \\&
    =
    \sqrt{\mathbb E[\Vert\sum\limits_{c=1}^N\varrho_c\sum_{j\in\mathcal S_c}\rho_{j,c}({\mathbf w}_j^{(t_{k})}-\bar{\mathbf v}_c^{(t_{k})})\Vert^2]}+\Vert\bar{\mathbf v}^{(t_{k})}-\mathbf w^*\Vert
    \\&
    \leq
    \sqrt{\sum\limits_{c=1}^N\varrho_c\sum_{j\in\mathcal S_c}\rho_{j,c}\mathbb E[\Vert{\mathbf w}_j^{(t_{k})}-\bar{\mathbf v}_c^{(t_{k})}\Vert^2]}+\Vert\bar{\mathbf v}^{(t_{k})}-\mathbf w^*\Vert
    = e_1^{(t_k)}+e_3^{(t_k)},
\end{align}
and therefore
\begin{align}\label{eq:final}
     \mathbb E[\Vert\bar{\mathbf w}^{(t_k)}-\mathbf w^*\Vert^2]\leq (e_1^{(t_k)}+e_3^{(t_k)})^2\leq 2(e_1^{(t_k)})^2+2(e_3^{(t_k)})^2.
\end{align}
We now show by induction that $e_1^{(t_k)}\leq Y_1\sqrt{\eta_k}$, $e_2^{(t_k)}\leq Y_2\eta_k$ and $e_3^{(t_k)}\leq Y_3\eta_k$ with $Y_1$, $Y_2$
and $Y_3$ defined in~\eqref{eq:Y1},~\eqref{eq:Y2} and~\eqref{eq:Y3}. The conditions trivially holds at the beginning of training at $k=0$ since $Y_1\geq0$, $Y_2\geq0$ and $Y_3\geq\eta_{\mathrm{max}}e_3^{(0)}$.
Now, assume $e_1^{(t_k)}\leq Y_1\sqrt{\eta_k}$, $e_2^{(t_k)}\leq Y_2\eta_k$ and $e_3^{(t_k)}\leq Y_3\eta_k$ for a certain $k\geq0$. We prove the condition holds for $k+1$ as well. 

To show $e_1^{(t_{k+1})}\leq\sqrt{\eta_{k+1}}Y_1$, we use~\eqref{eq:e1_main} of Proposition~\ref{lem:main_gap} and the induction hypothesis ($e_1^{(t_{k})}\leq\sqrt{\eta_{k}}Y_1$), yielding the sufficient condition
 \begin{align} 
&\left(1-\eta_k/\eta_{\mathrm{max}}C_1\right)\eta_{k}Y_1^2
    +\eta_{k}^2(\tau-(1-\alpha)\Delta)(\sigma^2+\phi^2)
    -\eta_{k+1}Y_1^2\leq 0.
\end{align}
Using the expression of $\eta_k=\frac{\eta_{\mathrm{max}}}{1+\gamma k}$, the above condition is equivalent to
$$
     -[C_1-\gamma]Y_1^2
    +\eta_{\mathrm{max}}(\tau-(1-\alpha)\Delta)(\sigma^2+\phi^2)
    -\gamma Y_1^2\frac{\gamma}{1+\gamma (k+1)}
    \leq 0.
$$
To satisfy the condition for all $k\geq 0$, the above condition is equivalent to
$$
\eta_{\mathrm{max}}(\tau-(1-\alpha)\Delta)(\sigma^2+\phi^2)
\leq 
[C_1-\gamma]Y_1^2,
$$
which is indeed verified since 
$\gamma<1-(1-\mu\eta_{\mathrm{max}})^{2(\tau-\Delta)}\leq C_1$
 and $Y_1^2=\frac{(\tau-(1-\alpha)\Delta)(\sigma^2+\phi^2)\eta_{\mathrm{max}}}{C_1-\gamma}.$
This completes the induction for $e_1^{(t_k)}$, showing that $(e_1^{(t_k)})^2\leq\eta_{k}Y_1^2,~\forall k$. 

To show $e_2^{(t_{k+1})}\leq\eta_{k+1}Y_2$, we use~\eqref{eq:x1_syn_inter2} of Proposition~\ref{lem:main_gap} and the induction hypothesis ($e_2^{(t_{k})}\leq\eta_{k}Y_2$), yielding the sufficient condition
 \begin{align} 
&\alpha(1+\lambda_+)^{\tau}\eta_k Y_2
    +\alpha 2\omega C_2\eta_k^2 Y_3
    +\alpha K_1\eta_k\delta
    -\eta_{k+1}Y_2\leq 0,
\end{align} 
Using the expression of $\eta_k=\frac{\eta_{\mathrm{max}}}{1+\gamma k}$, the above condition can be written as:
\begin{align}\label{eq:e2_finTransform}
    & \frac{\alpha(1+\lambda_+)^{\tau} Y_2}{\eta_{\mathrm{max}}}
    +\alpha 2\omega C_2Y_3
    +\frac{\alpha K_1\delta}{\eta_{\mathrm{max}}}
    - \frac{Y_2}{\eta_{\mathrm{max}}(1+\gamma)}
    - \frac{Y_2\gamma^2 k}{\eta_{\mathrm{max}}(1+\gamma(k+1))(1+\gamma)}
    - \frac{\alpha 2\omega C_2Y_3\gamma k}{1+\gamma k}
    \leq 0,
\end{align}
To satisfy the condition for all $k\geq 0$, the above condition is equivalent to
 \begin{align}\label{eq:e2_cond2}
    & \frac{[1-\alpha(1+\gamma)(1+\lambda_+)^{\tau}]Y_2-\alpha K_1\delta(1+\gamma)}{\eta_{\mathrm{max}}}
    -\alpha 2\omega C_2Y_3(1+\gamma)
    \geq 0.
 \end{align}

To show $e_3^{(t_{k+1})}\leq\eta_{k+1}Y_3$, we use~\eqref{eq:e3_sync} of Proposition~\ref{lem:main_gap} and the induction hypothesis ($e_3^{(t_{k})}\leq\eta_{k}Y_3$), yielding the sufficient condition
\begin{align}
    (1-\eta_k\beta C_3)Y_3\eta_k
    +[C_2Y_2+K_2\delta] \eta_k^2
    - Y_3\eta_{k+1}\leq 0.
\end{align}
Using the expression of $\eta_k=\frac{\eta_{\mathrm{max}}}{1+\gamma k}$, the above condition can be written as:
\begin{align}
Y_3[\gamma -\eta_{\max}\beta C_3]
    +[C_2Y_2+K_2\delta]\eta_{\max}
 -Y_3\gamma^2(1+\gamma k+\gamma)\leq 0.
\end{align}
To satisfy the condition for all $k\geq 0$, the above condition is equivalent to
\begin{align} \label{eq:cond_Y3}
    Y_3[\gamma -\eta_{\max}\beta C_3]
    +[C_2Y_2+K_2\delta]\eta_{\max}
    \leq 0.
\end{align}
 To show $e_2^{(t_k)}\leq\eta_{k+1}Y_2$ and $e_3^{(t_k)}\leq\eta_{k+1}Y_3$, the conditions $e_3^{(t_k)}\geq \eta_{\mathrm{max}}e_3^{(0)}$,~\eqref{eq:e2_cond2} and~\eqref{eq:cond_Y3} need to be satisfied simultaneously. To satisfy this, we need $\gamma<C_3\eta_{\mathrm{max}}\beta$ and
 \begin{align}
     Y_3\geq \eta_{\mathrm{max}}e_3^{(0)},
 \end{align}
 \begin{align}
     Y_3\leq
    \frac{[1-\alpha(1+\gamma)(1+\lambda_+)^{\tau}]Y_2-\alpha K_1\delta(1+\gamma)}{\eta_{\mathrm{max}}\alpha 2\omega C_2(1+\gamma)},
 \end{align}
 and
 \begin{align}
     Y_3\geq\frac{[C_2Y_2+K_2\delta]\eta_{\max}}{\eta_{\max}\beta C_3-\gamma}.
 \end{align}
 Using the definition of $Y_3$ in~\eqref{eq:Y3}, the conditions above become equivalent to 
 \begin{align}
     Y_3\leq
    \frac{[1-\alpha(1+\gamma)(1+\lambda_+)^{\tau}]Y_2-\alpha K_1\delta(1+\gamma)}{\eta_{\mathrm{max}}\alpha 2\omega C_2(1+\gamma)},
 \end{align}
 yielding the sufficient conditions
 \begin{align}
     \eta_{\mathrm{max}}e_3^{(0)}\leq
    \frac{[1-\alpha(1+\gamma)(1+\lambda_+)^{\tau}]Y_2-\alpha K_1\delta(1+\gamma)}{\eta_{\mathrm{max}}\alpha 2\omega C_2(1+\gamma)},
 \end{align}
 and 
 \begin{align}
     \frac{[C_2Y_2+K_2\delta]\eta_{\max}}{\eta_{\max}\beta C_3-\gamma}\leq
    \frac{[1-\alpha(1+\gamma)(1+\lambda_+)^{\tau}]Y_2-\alpha K_1\delta(1+\gamma)}{\eta_{\mathrm{max}}\alpha 2\omega C_2(1+\gamma)},
 \end{align}
which can be verified since $\alpha<\alpha^*$ together with the definition of $Y_2$ given in~\eqref{eq:Y2}.
This completes the induction showing that $e_2^{(t_k)}\leq\eta_{k+1}Y_2$ and $e_3^{(t_k)}\leq\eta_{k+1}Y_3$. Finally, applying the result of induction for $e_1^{(t_k)}, e_2^{(t_k)}$ and $e_3^{(t_k)}$ into~\eqref{eq:final} completes the proof.
\end{proof}

\pagebreak
\section{Proof of Proposition~\ref{lem:main_gap}} \label{app:main_gap} 
\begin{proposition} \label{lem:main_gap}
    Under Assumptions \ref{beta},~\ref{assump:SGD_noise} and~\ref{assump:sub_err}, if $\eta_{k}=\frac{\eta_{\mathrm{max}}}{1+\gamma k}$, where $\eta_{\mathrm{max}}<\min\left\{\frac{2}{\beta+\mu},\frac{(\tau-\Delta)\mu}{\beta^2[(1+\lambda_+)^{\tau}-1-\tau\lambda_+]}\right\}$, using {\tt DFL} for ML model training, $(e_1^{(t_{k+1})})^2$, $e_2^{(t_{k+1})}$ and $e_3^{(t_{k+1})}$ across global synchronizations can be bounded as 
\begin{align} \label{eq:e1_main}
    (e_1^{(t_{k+1})})^2
    \leq&
    \left(1-\eta_k/\eta_{\mathrm{max}}C_1\right)(e_1^{(t_k)})^2
    +\eta_{k}^2 (\tau-(1-\alpha)\Delta)(\sigma^2+\phi^2),
\end{align} 
\begin{align} \label{eq:x1_syn_inter2}
    e_2^{(t_{k+1})}&\leq 
    \alpha(1+\lambda_+)^{\tau}e_2^{(t_k)}
    +\eta_k\alpha 2\omega C_2e_3^{(t_k)}
    +\eta_k\alpha K_1\delta,
\end{align}
\begin{align} \label{eq:e3_sync}
     &  e_3^{(t_{k+1})}\leq
    (1-\eta_k\beta C_3) e_3^{(t_k)}
    +C_2 \eta_ke_2^{(t_k)}
    +\eta_k^2 K_2\delta,
\end{align}
 where 
 \begin{align}\label{eq:K1}
     C_1 \triangleq 1-((1-\alpha)(1-\mu\eta_{\mathrm{max}})^{2(\tau-\Delta)}+\alpha(1-\mu\eta_{\mathrm{max}})^{2\tau}),
 \end{align}
  \begin{align}\label{eq:C2}
     C_2\triangleq\frac{2\beta}{\sqrt{8\omega+1}}[(1+\lambda_+)^{\tau}-1],
 \end{align}
 \begin{align}\label{eq:C1}
     &C_3 \triangleq (\tau-\Delta)\mu/\beta 
        - \eta_{\mathrm{max}}\beta[(1+\lambda_+)^{\tau}-1-\tau\lambda_+],
 \end{align}
 \begin{align}\label{eq:K2}
     K_1 \triangleq \frac{\mu}{-\beta\lambda_+\lambda_-}[(1+\lambda_+)^{\tau}-1],
 \end{align}
 \begin{align}\label{eq:C3}
     K_2\triangleq \frac{\beta}{\sqrt{1+8\omega}}\sum_{\ell=0}^{\tau-2}\left(\begin{array}{c}\tau\\\ell+2\end{array}\right)[\lambda_+^{\ell+1}-\lambda_-^{\ell+1}],
 \end{align}
 with $\lambda_\pm$ defined in~\eqref{eq:eign+-} of Lemma~\ref{lem:main}.
\end{proposition}
\begin{proof}
We prove this result by further upper bounding $e_1^{(t_{k+1})}$ in Lemma~\ref{lem:main}. Therein, we found that
\begin{align} \label{eq:e1_lem2}
        &(e_1^{(t_{k+1})})^2
        \leq [(1-\alpha)(1-\mu\eta_{k})^{2(\tau-\Delta)}+\alpha(1-\mu\eta_{k})^{2\tau}](e_1^{(t_k)})^2
        +\eta_{k}^2(\tau-(1-\alpha)\Delta)(\sigma^2+\phi^2).
    \end{align} 
To bound $(e_1^{(t_{k+1})})^2$ in~\eqref{eq:e1_lem2}, we
 use the fact that $[(1-\alpha)(1-\mu\eta_{k})^{2(\tau-\Delta)}+\alpha(1-\mu\eta_{k})^{2\tau}]$ is a convex function of $\eta_k\in[0,\eta_{\mathrm{max}}]$ (in fact, $\mu\eta_{\mathrm{max}}\leq 1$ since $\eta_{\mathrm{max}}\leq\frac{2}{\beta+\mu}$), hence
\begin{align} \label{eq:cvx_eta}
    &(1-\alpha)(1-\mu\eta_{k})^{2(\tau-\Delta)}+\alpha(1-\mu\eta_{k})^{2\tau}
    \leq
    1-\eta_k/\eta_{\mathrm{max}}C_1.
\end{align}
 Applying the result from~\eqref{eq:cvx_eta} into~\eqref{eq:e1_lem2}, gives us the result in~\eqref{eq:e1_main}.
Similarly for $e_2^{(t_{k+1})}$, we found
  \begin{align} 
    e_2^{(t_{k+1})}&\leq 
    \alpha\Pi_{+,t_{k+1}}e_2^{(t_k)}
    +\alpha\frac{4\omega}{\sqrt{8\omega+1}}[\Pi_{+,t_{k+1}}-1]e_3^{(t_k)}
+\alpha\frac{\mu}{-\beta^2\lambda_+\lambda_-}(\Pi_{+,t_{k+1}}-1)\delta,
    \end{align}
    where $\Pi_{\{+,-\},t}=[1+\eta_{k}\beta\lambda_{\{+,-\}}]^{t-t_{k}}$.
    Using convexity of $\Pi_{+,t_{k+1}}$ in $\eta_k\beta\in[0,1]$, we bound it as $\Pi_{+,t_{k+1}}\leq1+ \eta_k\beta [(1+\lambda_+)^{\tau}-1]$. Applying it into the above inequality yields
    \begin{align}
    e_2^{(t_{k+1})}&\leq 
    \alpha(1+\lambda_+)^{\tau}e_2^{(t_k)}
    +\eta_{k}\alpha 2\omega C_2e_3^{(t_k)}
    +\eta_{k}\alpha K_1\delta.
    \end{align}
    Finally, we found in~\eqref{eq:wc-w*_sync1} of Lemma~\ref{lem:main} that
    \begin{align} \label{eq:e3_init}
         &e_3^{(t_{k+1})}\leq
        \Psi_1(\eta_k) e_3^{(t_k)}
          \nonumber \\&
        +\underbrace{2g_{3}[(1-\alpha)\Pi_{+,t_{k+1}-\Delta}+\alpha\Pi_{+,t_{k+1}}-1]}_{(a)}e_2^{(t_k)}
        \nonumber\\&
        +\underbrace{\left[(1-\alpha)[g_{5}(\Pi_{+,t_{k+1}-\Delta}-1)+g_{6}(\Pi_{-,t_{k+1}-\Delta}-1)]
        +\alpha[g_{5}(\Pi_{+,t_{k+1}}-1)+g_{6}(\Pi_{-,t_{k+1}}-1)]\right]}_{(b)}\delta/\beta,
    \end{align}
    where $\Psi_1(\eta_k)$, $g_3$, $g_5$ and $g_6$ is defined in~\eqref{eq:Psi_1},~\eqref{eq:g3},~\eqref{eq:g5} and~\eqref{eq:g6} of Lemma~\ref{lem:main}. We bound $\Psi_1(\eta_k)$ as follows.
    Applying the binomial expansion, we have
    \begin{align} \label{eq:phi_bi}
        &\frac{g_{1}\Pi_{+,t_{k}+\ell}+g_{2}\Pi_{-,t_{k}+\ell}-1}{\eta_k\beta}
        \nonumber \\&
        =-\ell\mu/\beta+\eta_k\beta\sum\limits_{r=2}^\ell\frac{\ell!}{r!(\ell-r)!}(\eta_k\beta)^{r-2}
        \left[\frac{1}{2}(1-\frac{1}{\sqrt{8\omega+1}})\lambda_+^r+\frac{1}{2}(1+\frac{1}{\sqrt{8\omega+1}})\lambda_-^r\right].
    \end{align}
    Since $\lambda_-\leq\lambda_+$, $\eta_k\leq\eta_{\mathrm{max}}$
    and $\eta_k\beta\leq1$, we can further upper bound~\eqref{eq:phi_bi} with
    \begin{align}
        &\frac{g_{1}\Pi_{+,t_{k}+\ell}+g_{2}\Pi_{-,t_{k}+\ell}-1}{\eta_k\beta}
        \leq -\ell\mu/\beta +\eta_k\beta\sum\limits_{r=2}^\ell\frac{\ell!}{r!(\ell-r)!}(\eta_k\beta)^{r-2}\lambda_+^r
        \nonumber \\&
        \leq -\ell\mu/\beta +\eta_{\mathrm{max}}\beta\sum\limits_{r=2}^\ell\frac{\ell!}{r!(\ell-r)!}\lambda_+^r
        \nonumber \\&
        \overset{(a)}{=} -\ell\mu/\beta +\eta_{\mathrm{max}}\beta[(1+\lambda_+)^\ell-1-\ell\lambda_+],
    \end{align}
    where $(a)$ comes from applying the binomial theorem.
    Note that $(1+\lambda_+)^\ell-1-\ell\lambda_+\geq0,~\forall \ell\geq0$. 
    Combining this result into~\eqref{eq:Psi_1} of Lemma~\ref{lem:main}, it follows that 
    \begin{align}
        &\frac{\Psi_1(\eta_k)-1}{\eta_k\beta}
        \leq 
        -(\tau-(1-\alpha)\Delta)\mu/\beta 
        \nonumber \\&
        + \eta_{\mathrm{max}}\beta[(1-\alpha)(1+\lambda_+)^{\tau-\Delta}+\alpha(1+\lambda_+)^{\tau}-1-(\tau-(1-\alpha)\Delta)\lambda_+]
        \nonumber \\&
        \leq 
        -(\tau-\Delta)\mu/\beta 
        + \eta_{\mathrm{max}}\beta[(1+\lambda_+)^{\tau}-1-\tau\lambda_+]
        \triangleq -C_3,
    \end{align}
    where in the last inequality comes from $\tau-(1-\alpha)\Delta\geq\tau-\Delta$ and $(1-\alpha)(1+\lambda_+)^{\tau-\Delta}+\alpha(1+\lambda_+)^{\tau}-1-(\tau-(1-\alpha)\Delta)\lambda_+\leq(1+\lambda_+)^{\tau}-1-\tau\lambda_+$.
    Therefore, under $\eta_{\mathrm{max}}<\frac{(\tau-\Delta)\mu}{\beta^2[(1+\lambda_+)^{\tau}-1-\tau\lambda_+]}$, we have $C_3>0$ and
    \begin{align}
        \Psi_1(\eta_k)\leq 1-\eta_k\beta C_3<1. 
    \end{align}
    Next, we bound $(a)$ in~\eqref{eq:e3_init}.
    Convexity of $\Pi_{+,t}-1$ with respect to $\eta_k\beta$ and $\eta_{\mathrm{max}}\beta\leq1$ implies that
    \begin{align}
        &2g_3[(1-\alpha)\Pi_{+,t_{k+1}-\Delta}+\alpha\Pi_{+,t_{k+1}}-1]
        \nonumber \\&
        \leq
        \eta_{k} \frac{2\beta}{\sqrt{8\omega+1}}[(1-\alpha)(1+\lambda_+)^{\tau-\Delta}+\alpha(1+\lambda_+)^{\tau}-1]
        \leq 
        \eta_k C_2,
    \end{align}
    where $g_3$ is defined in~\eqref{eq:g3} of Lemma~\ref{lem:main},
    with $$C_2= \frac{2\beta}{\sqrt{8\omega+1}}[(1+\lambda_+)^{\tau}-1].$$
Finally, we bound $(b)$ in~\eqref{eq:e3_init}, using the binomial expansion and the expressions of $g_5$ and $g_6$ 
    $$
    g_{5}(\Pi_{+,t}-1)+g_{6}(\Pi_{-,t}-1)
    =(\eta_k\beta)^2\frac{1}{\sqrt{1+8\omega}}\sum_{\ell=0}^{t-t_k-2}\left(\begin{array}{c}t-t_k\\\ell+2\end{array}\right)(\eta_k\beta)^{\ell}[\lambda_+^{\ell+1}-\lambda_-^{\ell+1}]
    \leq \eta_k^2\beta K_2.
    $$
Using these bounds in~\eqref{eq:e3_init} yield the final result in~\eqref{eq:e3_sync}.
\end{proof} 
\pagebreak

\section{Lemmas and Auxiliary Results}\label{app:lemmas}
To improve the tractability of the proofs, we provide a set of lemmas in the following, which will be used to obtain the main results of the paper.

\begin{lemma} \label{lem:An_oneSTP}
For $t\in\mathcal T_k$ before performing global synchronization, under Assumptions \ref{beta},~\ref{assump:SGD_noise} and~\ref{assump:sub_err}, if $\eta_{k}\leq\frac{2}{\mu+\beta},~\forall k$, using {\tt DFL} for ML model training, in $t\in\mathcal T_k$, the one-step behaviors of $(e_1^{(t+1)})^2$, $e_2^{(t+1)}$ and $e_3^{(t+1)}$ are presented as follows:
\begin{align} \label{eq:e1_oneSTP}
    &(e_1^{(t+1)})^2
    \leq
    (1-\mu\eta_{k})^2(e_1^{(t)})^2
    +\eta_{k}^2(\sigma^2+\phi^2),
    \\\label{eq:e2_oneSTP}
    &e_2^{(t+1)}\leq
    (1+\eta_{k}(\beta-\mu))e_2^{(t)} 
    +2\omega\eta_{k}\beta e_3^{(t)}
    +\eta_{k}\delta,
\\ \label{eq:e3_oneSTP}
&   e_3^{(t+1)} \leq
     (1-\eta_{k}\mu)e_3^{(t)} 
    +\eta_{k}\beta e_2^{(t)}.
\end{align} 
\end{lemma}

\begin{proof}
To bound $e_1^{(t)}$, we first use the definition of $\mathbf w_i^{(t+1)},~\forall i \in\mathcal S_c$ in \eqref{eq:w_i-gen} and $\mathbf v_c^{(t+1)}$ in \eqref{eq:v_c} to get, 
\begin{align} \label{eq:e_tmp1}
    &\mathbf w_i^{(t+1)} - \bar{\mathbf v}_c^{(t+1)}
    =(1-\Theta_c^{(t)})(\mathbf w_i^{(t)}-\eta_{k} \nabla F_i({\mathbf w}_i^{(t)})-\bar{\mathbf v}_c^{(t)}+\eta_{k} \nabla\bar F_c(\bar{\mathbf v}_c^{(t)}))
    \nonumber \\&
    +\Theta_c^{(t)}(\bar{\mathbf w}_c^{(t)}-\eta_{k}\sum\limits_{j\in\mathcal S_{c}}\rho_{j,c}\nabla F_j({\mathbf w}_j ^{(t)})-\bar{\mathbf v}_c^{(t)}+\eta_{k} \nabla\bar F_c(\bar{\mathbf v}_c^{(t)}))
    \nonumber \\&
    -\eta_{k}(1-\Theta_c^{(t)})\mathbf n_{i}^{(t)}-\eta_{k}\Theta_c^{(t)}\sum\limits_{j\in\mathcal S_c}\rho_{j,c}\mathbf n_{j}^{(t)}.
\end{align}
Then,
\begin{align}\label{eq:A_new}
    &(e_1^{(t+1)})^2\triangleq\mathbb E\Big[\sum\limits_{c=1}^N\varrho_c\sum\limits_{j\in\mathcal S_c}\rho_{j,c}\Vert\mathbf w_j^{(t+1)} - \bar{\mathbf v}_c^{(t+1)}\Vert^2\Big]
    \nonumber \\&
    \leq
    \mathbb E\Big[\sum\limits_{c=1}^N\varrho_c(1-\Theta_c^{(t)})\sum\limits_{j\in\mathcal S_c}\rho_{j,c}\Vert\mathbf w_j^{(t)}-\eta_{k} \nabla F_j({\mathbf w}_j^{(t)})-\bar{\mathbf v}_c^{(t)}+\eta_{k} \nabla\bar F_c(\bar{\mathbf v}_c^{(t)})\Vert^2\Big]
    \nonumber \\&
    +\mathbb E\Big[\sum\limits_{c=1}^N\varrho_c\Theta_c^{(t)}\Vert\bar{\mathbf w}_c^{(t)}-\eta_{k}\sum\limits_{j\in\mathcal S_{c}}\rho_{j,c}\nabla F_j({\mathbf w}_j ^{(t)})-\bar{\mathbf v}_c^{(t)}+\eta_{k} \nabla\bar F_c(\bar{\mathbf v}_c^{(t)})\Vert^2\Big]+\eta_{k}^2\sigma^2
        \nonumber \\&
    \leq
    \mathbb E\Big[\sum\limits_{c=1}^N\varrho_c(1-\Theta_c^{(t)})\sum\limits_{j\in\mathcal S_c}\rho_{j,c}\Vert\mathbf w_j^{(t)}-\bar{\mathbf v}_c^{(t)}-\eta_{k} (\nabla F_j({\mathbf w}_j^{(t)})- \nabla\bar F_c(\bar{\mathbf v}_c^{(t)}))\Vert^2\Big]
    \nonumber \\&
    +\mathbb E\Big[\sum\limits_{c=1}^N\varrho_c\Theta_c^{(t)}
    \sum\limits_{j\in\mathcal S_{c}}\rho_{j,c}
    \Vert\mathbf w_j^{(t)}-\eta_{k}\nabla F_j({\mathbf w}_j ^{(t)})-\bar{\mathbf v}_c^{(t)}+\eta_{k} \nabla F_j(\bar{\mathbf v}_c^{(t)})\Vert^2\Big]+\eta_{k}^2\sigma^2,
\end{align}
where the last step
follows from
$\sum\limits_{j\in\mathcal S_c}\rho_{j,c}
 F_j(\bar{\mathbf v}_c^{(t)})
=
\bar F_c(\bar{\mathbf v}_c^{(t)}),
$
$\sum\limits_{j\in\mathcal S_c}\rho_{j,c}
{\mathbf w}_j^{(t)}
=\bar{\mathbf w}_c^{(t)}
$
and convexity of $\Vert\cdot\Vert^2$. Using again the fact that
$\sum\limits_{j\in\mathcal S_c}\rho_{j,c}
 F_j(\bar{\mathbf v}_c^{(t)})
=
\bar F_c(\bar{\mathbf v}_c^{(t)}),
$
we further bound
\begin{align}
    &\sum\limits_{j\in\mathcal S_c}\rho_{j,c}\Vert\mathbf w_j^{(t)}-\bar{\mathbf v}_c^{(t)}-\eta_{k} (\nabla F_j({\mathbf w}_j^{(t)})- \nabla\bar F_c(\bar{\mathbf v}_c^{(t)}))\Vert^2
    \nonumber \\&
    =
\sum\limits_{j\in\mathcal S_c}\rho_{j,c}\Vert\mathbf w_j^{(t)}-\bar{\mathbf v}_c^{(t)}-\eta_{k}(\nabla F_j({\mathbf w}_j^{(t)})-\nabla F_j(\bar{\mathbf v}_c^{(t)})) -\eta_{k}\left(\nabla F_j(\bar{\mathbf v}_c^{(t)})- \nabla\bar F_c(\bar{\mathbf v}_c^{(t)})\right)\Vert^2
    \nonumber \\&
    \leq
\sum\limits_{j\in\mathcal S_c}\rho_{j,c}\Vert\mathbf w_j^{(t)}-\bar{\mathbf v}_c^{(t)}-\eta_{k} (\nabla F_j({\mathbf w}_j^{(t)})-\eta_{k} \nabla F_j(\bar{\mathbf v}_c^{(t)}))\Vert^2
    +\eta_{k}^2\sum\limits_{j\in\mathcal S_c}\rho_{j,c}\Vert\nabla F_j(\bar{\mathbf v}_c^{(t)})- \nabla\bar F_c(\bar{\mathbf v}_c^{(t)})\Vert^2.
    \nonumber
\end{align}
Furthermore, 
$\Vert\nabla F_j(\bar{\mathbf v}_c^{(t)})- \nabla\bar F_c(\bar{\mathbf v}_c^{(t)})\Vert^2
\leq
(\delta_c+2\omega_c\beta\Vert\bar{\mathbf v}_c^{(t)}-\mathbf w^*\Vert)^2
\leq
2\delta_c^2+4\omega_c\beta\Vert\bar{\mathbf v}_c^{(t)}-\mathbf w^*\Vert^2
$ (Definition~\ref{gradDiv_c}).
Combining these bounds together into~\eqref{eq:A_new} and using Fact~\ref{fact:3} yields
\begin{align}\label{eq:A_new}
    &(e_1^{(t+1)})^2
    \leq
    (1-\mu\eta_k)^2\mathbb E\Big[\sum\limits_{c=1}^N\varrho_c
    \sum\limits_{j\in\mathcal S_{c}}\rho_{j,c}
    \Vert\bar{\mathbf w}_j^{(t)}-\bar{\mathbf v}_c^{(t)}\Vert^2\Big]
    \nonumber \\&
    +\eta_{k}^2
    \sum\limits_{c=1}^N\varrho_c(1-\Theta_c^{(t)})(2\delta_c^2+4\omega_c^2\beta^2\Vert\bar{\mathbf v}_c^{(t)}-\mathbf w^*\Vert^2)
    +\eta_{k}^2\sigma^2.
\end{align}
Assuming that $\Theta_c^{(t)}$ is chosen such that $\sum\limits_{c=1}^N\varrho_c(1-\Theta_c^{(t)})(2\delta_c^2+4\omega_c^2\beta^2\Vert\bar{\mathbf v}_c^{(t)}-\mathbf w^*\Vert^2)\leq\phi^2$, (this will be part of the control algorithm, see Assumption~\ref{assump:sub_err}) we can further upper bound~\eqref{eq:A_new} and obtain the result in~\eqref{eq:e1_oneSTP}.

Next, we  bound $e_2$.
Using~\eqref{eq:v_c} we find that
    \begin{align} \label{eq:w-_3}
        &\bar{\mathbf v}^{(t+1)}=
        \bar{\mathbf v}^{(t)}
        -\eta_{k}\sum\limits_{d=1}^N\varrho_{d}\nabla \bar{F}_d(\bar{\mathbf v}_d^{(t)}).
    \end{align}
It then follows, after algebraic manipulations,
    \begin{align} 
        &\bar{\mathbf v}_c^{(t+1)}-\bar{\mathbf v}^{(t+1)}=\bar{\mathbf v}_c^{(t)}-\bar{\mathbf v}^{(t)}
        -\eta_{k}\Big(\nabla\bar F_c(\bar{\mathbf v}_c^{(t)})-\nabla\bar F_c(\bar{\mathbf v}^{(t)})\Big)
        \nonumber \\&
        +\eta_{k}\sum\limits_{d=1}^N\varrho_{d}\Big(\nabla\bar F_d(\bar{\mathbf v}_d^{(t)})-\nabla\bar F_d(\bar{\mathbf v}^{(t)})\Big)
        -\eta_{k}\Big(\nabla\bar F_c(\bar{\mathbf v}^{(t)})-\nabla F(\bar{\mathbf v}^{(t)})\Big).
    \end{align}   
    Taking the norm-2 of both hand sides of the above equality and applying the triangle inequality results in
    \begin{align} \label{eq:tri_wc}
        &\Vert\bar{\mathbf v}_c^{(t+1)}-\bar{\mathbf v}^{(t+1)}\Vert\leq
        \left\Vert\bar{\mathbf v}_c^{(t)}-\bar{\mathbf v}^{(t)} -\eta_{k}[\nabla\bar F_c(\bar{\mathbf v}_c^{(t)})-\nabla\bar F_c(\bar{\mathbf v}^{(t)})]\right\Vert
        \nonumber \\&
        +\eta_{k}\sum\limits_{d=1}^N\varrho_{d}\Vert\nabla\bar F_d(\bar{\mathbf v}_d^{(t)})-\nabla\bar F_d(\bar{\mathbf v}^{(t)}))\Vert
        +\eta_{k}\Vert\nabla\bar F_c(\bar{\mathbf v}^{(t)})-\nabla F(\bar{\mathbf v}^{(t)})\Vert.
    \end{align}   
    Using $\beta$-smoothness of $F_i(\cdot),\forall i$ (hence of $\bar F_d(\cdot)$), Definition \ref{gradDiv}, Fact~\ref{fact:3}, and adding over $\sum_c\rho_c$,
     we further bound the right hand side of~\eqref{eq:tri_wc} as
\begin{align} \label{eq:tri_wc2_3}
    &e_2^{(t+1)}\triangleq\sum\limits_{c=1}^N\varrho_{c}\Vert\bar{\mathbf v}_c^{(t+1)}-\bar{\mathbf v}^{(t+1)}\Vert\leq
    (1+\eta_{k}(\beta-\mu))\sum\limits_{c=1}^N\varrho_{c}\Vert\bar{\mathbf v}_c^{(t)}-\bar{\mathbf v}^{(t)}\Vert
    +2\omega\eta_{k}\beta\Vert\bar{\mathbf v}^{(t)}-\mathbf w^*\Vert
    +\eta_{k}\delta,
\end{align}   
which proves~\eqref{eq:e1_oneSTP}.
Finally, we  bound $e_3$. From \eqref{eq:w-_3}, we get 
\begin{align} 
        &\bar{\mathbf v}^{(t+1)}-\mathbf w^* =~ \bar{\mathbf v}^{(t)}-\mathbf w^*-\eta_{k} \nabla F(\bar{\mathbf v}^{(t)})
        -\eta_{k} \sum\limits_{c=1}^N\varrho_{d} [\nabla \bar F_d(\bar{\mathbf v}_d^{(t)})-\nabla \bar F_d(\bar{\mathbf v}^{(t)})].
    \end{align}
    Taking the norm of both hand sides of the above equality and applying the triangle inequality gives us
    \begin{align} \label{29_3}
        &\Vert\bar{\mathbf v}^{(t+1)}-\mathbf w^*\Vert \leq \Vert\bar{\mathbf v}^{(t)}-\mathbf w^*-\eta_{k} \nabla F(\bar{\mathbf v}^{(t)})\Vert
        +\eta_{k} \sum\limits_{d=1}^N\varrho_{d} \Vert\nabla \bar F_d(\bar{\mathbf v}_d^{(t)})-\nabla \bar F_d(\bar{\mathbf v}^{(t)})\Vert.
    \end{align}
    Using $\beta$-smoothness of $F_i(\cdot)$ (hence of $\bar F_c(\cdot)$) and Fact~\ref{fact:3}, we further bound
    \begin{align}
        e_3^{(t+1)}\triangleq\Vert\bar{\mathbf v}^{(t+1)}-\mathbf w^*\Vert \leq& 
         (1-\eta_{k}\mu)
        \Vert\bar{\mathbf v}^{(t)}-\mathbf w^*\Vert
        +\eta_{k}\beta \sum\limits_{d=1}^N\varrho_{d} \Vert\bar{\mathbf v}_d^{(t)}-\bar{\mathbf v}^{(t)}\Vert,
    \end{align}
    yielding~\eqref{eq:e3_oneSTP}.
\end{proof}

\begin{lemma} \label{lem:main}
    Under Assumptions \ref{beta},~\ref{assump:SGD_noise} and~\ref{assump:sub_err}, if $\eta_{k}\leq\frac{2}{\beta+\mu},~\forall k$, using {\tt DFL} for ML model training, $e_1^{t_{k+1}}$, $e_2^{t_{k+1}}$ and $e_3^{(t_{k+1})}$ across global synchronization periods can be bounded as 
 \begin{align} \label{eq:e1_orig}
    &(e_1^{t_{k+1}})^2
    \leq [(1-\alpha)(1-\mu\eta_{k})^{2(\tau-\Delta)}+\alpha(1-\mu\eta_{k})^{2\tau}](e_1^{(t_k)})^2
    +[\tau-(1-\alpha)\Delta]\eta_{k}^2(\sigma^2+\phi^2),
\end{align} 
\begin{align} \label{eq:x1_syn_inter1}
    e_2^{(t_{k+1})}&\leq 
    \alpha\Pi_{+,t_{k+1}}e_2^{(t_k)}
    +\alpha\frac{4\omega}{\sqrt{8\omega+1}}[\Pi_{+,t_{k+1}}-1]e_3^{(t_k)}
    +\alpha\frac{\mu}{-\beta^2\lambda_+\lambda_-}[\Pi_{+,t_{k+1}}-1]\delta,
\end{align}
\begin{align} \label{eq:wc-w*_sync1}
    &e_3^{(t_{k+1})}\leq
    \Psi_1(\eta_k) e_3^{(t_k)}
    +2g_{3}[(1-\alpha)\Pi_{+,t_{k+1}-\Delta}+\alpha\Pi_{+,t_{k+1}}-1]e_2^{(t_k)}
    \nonumber\\&
    +\left[(1-\alpha)[g_{5}(\Pi_{+,t_{k+1}-\Delta}-1)+g_{6}(\Pi_{-,t_{k+1}-\Delta}-1)] 
    +\alpha[g_{5}(\Pi_{+,t_{k+1}}-1)+g_{6}(\Pi_{-,t_{k+1}}-1)]\right]\delta/\beta.
\end{align}
where 
\begin{align}\label{eq:Psi_1}
    \Psi_1(\eta_k)\triangleq (1-\alpha)[g_{1}\Pi_{+,t_{k+1}-\Delta}+g_{2}\Pi_{-,t_{k+1}-\Delta}]
    + \alpha[g_{1}\Pi_{+,t_{k+1}}+g_{2}\Pi_{-,t_{k+1}}]
\end{align}
and
\begin{align}\label{eq:eign+-}
    \lambda_{\pm} =\frac{1}{2}-\frac{\mu}{\beta}\pm\frac{\sqrt{8\omega+1}}{2},
\end{align}
with $\lambda_+>0$, $\lambda_-<0$, $\Pi_{\{+,-\},t}=[1+\eta_{k}\beta\lambda_{\{+,-\}}]^{t-t_{k}}$ and $g_1$, $g_2$, $g_3$, $g_5$, $g_6$ defined in~\eqref{eq:g1},~\eqref{eq:g2},~\eqref{eq:g3},~\eqref{eq:g5},~\eqref{eq:g6}.
\end{lemma}

\begin{proof}
\subsection{Obtaining the upper bound of $e_1^{(t_{k+1})}$ at global synchronization} 
Using the one-step dynamics in~\eqref{eq:e1_oneSTP}, Lemma~\ref{lem:An_oneSTP}, we find before global synchronization
\begin{align}
\label{e1beforesync}
    &(e_1^{(t)})^2
    \leq
    (1-\mu\eta_{k})^{2(t-t_k)}(e_1^{(t_k)})^2
    +\sum_{\ell=0}^{t-t_k-1}(1-\mu\eta_{k})^{2\ell}\eta_{k}^2(\sigma^2+\phi^2)
    \nonumber \\&
    \leq
    (1-\mu\eta_{k})^{2(t-t_k)}(e_1^{(t_k)})^2
    +(t-t_k)\eta_{k}^2(\sigma^2+\phi^2).
\end{align}
         Next, we obtain the behavior of $e_1^{(t)}$ at global synchronization by using the definition of $\mathbf w_i^{(t)}$, $\bar{\mathbf v}_c^{(t)}$ and the global synchronization scheme in~\eqref{eq:aggr_alpha} as follows:
\begin{align}
    &\mathbf w_i^{(t_{k+1})}-\bar{\mathbf v}_c^{(t_{k+1})}
    = (1-\alpha)\sum\limits_{d=1}^N\varrho_d\sum_{j\in\mathcal S_d}\rho_{j,d}({\mathbf w}_j^{(t_{k+1}-\Delta)}-\bar{\mathbf v}_d^{(t_{k+1}-\Delta)})
    +\alpha\left(\widetilde{\mathbf w}_i^{(t_{k+1})}-\widetilde{\mathbf v}_c^{(t_{k+1})}\right), 
\end{align}
where $\widetilde{\mathbf w}_i^{(t_{k+1})}$ and $\widetilde{\mathbf v}_c^{(t_{k+1})}$ are the local model and subnet noise-free variables right before global synchronization, as opposed to $\mathbf w_i^{(t_{k+1})}$ and $\mathbf v_c^{(t_{k+1})}$ defined right after global synchronization.
  Taking the squared norm on both hand sides of the above equality and applying Jensen's inequality (convexity of $\Vert\cdot\Vert^2$) yields
\begin{align} 
    &\Vert\mathbf w_i^{(t_{k+1})}-\bar{\mathbf v}_c^{(t_{k+1})}\Vert^2
    \leq (1-\alpha)\sum\limits_{d=1}^N\varrho_d\sum_{j\in\mathcal S_d}\rho_{j,d}\Vert{\mathbf w}_j^{(t_{k+1}-\Delta)}-\bar{\mathbf v}_d^{(t_{k+1}-\Delta)}\Vert^2
    +\alpha\Vert\widetilde{\mathbf w}_i^{(t_{k+1})}-\widetilde{\mathbf v}_c^{(t_{k+1})}\Vert^2.
\end{align} 
Therefore, 
\begin{align} \label{eq:e1_sy}
    &(e_1^{(t_{k+1})})^2=\sum\limits_{c=1}^N\varrho_c\sum_{i\in\mathcal S_c}\rho_{i,c}\Vert\mathbf w_i^{(t_{k+1})}-\bar{\mathbf v}_c^{(t_{k+1})}\Vert^2
        \nonumber \\&
    \leq (1-\alpha)\sum\limits_{c=1}^N\varrho_c\sum_{j\in\mathcal S_c}\rho_{j,c}\Vert{\mathbf w}_j^{(t_{k+1}-\Delta)}-\bar{\mathbf v}_c^{(t_{k+1}-\Delta)}\Vert^2
    +\alpha\sum\limits_{c=1}^N\varrho_c\sum_{j\in\mathcal S_c}\rho_{j,c}\left\Vert\widetilde{\mathbf w}_j^{(t_{k+1})}-\widetilde{\mathbf v}_c^{(t_{k+1})}\right\Vert^2.
\end{align} 
Note that the terms above are upper bounded by \eqref{e1beforesync} before global synchronization, hence they can be bounded as
$$
\sum\limits_{c=1}^N\varrho_c\sum_{j\in\mathcal S_c}\rho_{j,c}\Vert{\mathbf w}_j^{(t_{k+1}-\Delta)}-\bar{\mathbf v}_c^{(t_{k+1}-\Delta)}\Vert^2
    \leq
    (1-\mu\eta_{k})^{2(\tau-\Delta)}(e_1^{(t_k)})^2+[\tau-\Delta]\eta_{k}^2(\sigma^2+\phi^2),
$$
$$
 \sum\limits_{c=1}^N\varrho_c\sum_{j\in\mathcal S_c}\rho_{j,c}\left\Vert\widetilde{\mathbf w}_j^{(t_{k+1})}-\widetilde{\mathbf v}_c^{(t_{k+1})}\right\Vert^2
    \leq
    (1-\mu\eta_{k})^{2\tau}(e_1^{(t_k)})^2
    +\tau\eta_{k}^2(\sigma^2+\phi^2).
$$
Using these bounds in~\eqref{eq:e1_sy} yields~\eqref{eq:e1_orig}. 

\subsection{Solving the coupled dynamics between $e_2$ and $e_3$} 

Let $\mathbf x^{(t)}=
     \begin{bmatrix}
         e_2^{(t)} & e_3^{(t)}
     \end{bmatrix}^\top$, with $e_2$ and $e_3$ defined in~\eqref{eq:def_e2} and~\eqref{eq:def_e3}. Using the one-step dynamics found in Lemma~\ref{lem:main} for $t\in\mathcal T_k$, we find (here, the vector inequality is entry-wise)
    \begin{align} \label{69}
        \mathbf x^{(t+1)}
        &\leq  
        [\mathbf I+\eta_{k}\beta\mathbf B]\mathbf  x^{(t)}
        +\eta_{k}\beta\mathbf z, 
    \end{align}
    where
    $
\mathbf z=  
\mathbf e_1\delta/\beta
    $
    ,    
    $\mathbf B=\begin{bmatrix} 1-\frac{\mu}{\beta} & 2\omega
    \\ 1 & -\frac{\mu}{\beta}\end{bmatrix}$, $\mathbf e_1=[1,0]^\top$. We also define  $\mathbf e_2=[0,1]^\top$.
    We aim to derive an upper bound on $\mathbf x^{(t)}$ denoted by $\mathbf x^{(t)}\leq\bar{\mathbf x}^{(t)}$. Using the above inequality, such upper bound is given by the recursion
     \begin{align} \label{69_3}
 \bar{\mathbf x}^{(t+1)}
=  
        [\mathbf I+\eta_{k}\beta\mathbf B]\bar{\mathbf x}^{(t)}
        +\eta_{k}\beta\mathbf z, \ \forall t\in\mathcal T_k,
    \end{align}
 initialized as $\bar{\mathbf x}^{(t_k)}={\mathbf x}^{(t_k)}$.
 To solve the coupled dynamic, we first apply eigen-decomposition on $\mathbf B$ yielding
    $\mathbf B=\mathbf U\mathbf D\mathbf U^{-1}$, where
    $$\mathbf D=\begin{bmatrix} \lambda_+ & 0
    \\ 0 & \lambda_-\end{bmatrix},\ 
    \mathbf U=\begin{bmatrix} \frac{1}{2}(1+\sqrt{8\omega+1}) & -\frac{1}{2}(\sqrt{8\omega+1}-1)
    \\ 1 & 1\end{bmatrix},\ 
\mathbf U^{-1}=\frac{1}{
    \sqrt{8\omega+1}
    }\begin{bmatrix} 1 & \frac{1}{2}(\sqrt{8\omega+1}-1)
    \\ -1 &\frac{1}{2}(\sqrt{8\omega+1}+1)\end{bmatrix}$$
with eigenvalues given by~\eqref{eq:eign+-}.
Using this decomposition in~\eqref{69_3} yields by induction
    \begin{align} 
    \bar{\mathbf x}^{(t)}=
\mathbf U(\mathbf I+\eta_{k}\beta\mathbf D)^{t-t_{k}}
         \mathbf U^{-1}{\mathbf x}^{(t_k)}
+\mathbf U\left[(\mathbf I+\eta_{k}\beta\mathbf D)^{t-t_{k}}-\mathbf I\right]
         \mathbf D^{-1}\mathbf U^{-1}\mathbf z.
    \end{align}
Therefore,
\begin{align} \label{eq:x2_dyn}
    e_2^{(t)}&=\mathbf e_1^\top\mathbf x^{(t)}\leq 
    \mathbf e_1^\top\bar{\mathbf x}^{(t)}
   \nonumber\\&=
  [m_{1}\Pi_{+,t}+m_{2}\Pi_{-,t}]e_3^{(t_k)}
  \nonumber \\&
+[m_{3}\Pi_{+,t}+m_{4}\Pi_{-,t}]e_2^{(t_k)}
\nonumber\\&
+[m_{5}(\Pi_{+,t}-1)+m_{6}(\Pi_{-,t}-1)]\delta/\beta, 
    \end{align}
    where we have defined $\Pi_{\{+,-\},t}=[1+\eta_{k}\beta\lambda_{\{+,-\}}]^{t-t_{k}}$ and constants $m_{1}$-$m_{8}$ as  
    \begin{flalign}
        m_{1}\triangleq
        &[\mathbf U]_{1,1}[\mathbf U^{-1}]_{1,2}
        =
        \frac{2\omega}{\sqrt{8\omega+1}},&
    \end{flalign}
    \begin{flalign}
        m_{2}\triangleq
        &[\mathbf U]_{1,2}[\mathbf U^{-1}]_{2,2}
        =
        -m_1, &
    \end{flalign}
    \begin{flalign}
        m_{3}\triangleq
         &[\mathbf U]_{1,1}[\mathbf U^{-1}]_{1,1}
         =\frac{\sqrt{8\omega+1}+1}{2\sqrt{8\omega+1}}, &
    \end{flalign}
    \begin{flalign}
        m_{4}\triangleq
        &[\mathbf U]_{1,2}[\mathbf U^{-1}]_{2,1}
        =1-m_3\geq 0, &
    \end{flalign}
\begin{flalign}
    m_{5}
    \triangleq
    &[\mathbf U]_{1,1}[\mathbf D^{-1}\mathbf U^{-1}]_{1,1}
    =
    \frac{\mu(\sqrt{8\omega+1}+1)+4\omega\beta}{-2\beta\lambda_+\lambda_-\sqrt{8\omega+1}}=\frac{\sqrt{8\omega+1}+1}{2\sqrt{8\omega+1}\lambda_+}\geq 0,&
\end{flalign}
\begin{flalign}
    m_{6}\triangleq
    &[\mathbf U]_{1,2}[\mathbf D^{-1}\mathbf U^{-1}]_{2,1}
    =
    \frac{\mu(\sqrt{8\omega+1}-1)-4\omega\beta}{-2\beta\lambda_+\lambda_-\sqrt{8\omega+1}}
    =\frac{\sqrt{8\omega+1}-1}{-2\sqrt{8\omega+1}\lambda_-}\leq 0.&
\end{flalign}
Since $\Pi_{-,t}\leq\Pi_{+,t}$, $\Pi_{+,t}-\Pi_{-,t}\leq 2[\Pi_{+,t}-1]$, 
 $m_3+m_4=1$ and $\Pi_{-,t}-1\geq-(\Pi_{+,t}-1)$, we can further upper bound
\begin{align} \label{eq:E_1}
    e_2^{(t)}&\leq 
    \frac{4\omega}{\sqrt{8\omega+1}}[\Pi_{+,t}-1]e_3^{(t_k)}
    +\Pi_{+,t}e_2^{(t_k)}
    +\frac{\mu}{-\beta^2\lambda_+\lambda_-}(\Pi_{+,t}-1)\delta
    . 
    \end{align}
Similarly, from the expression of $\bar{\mathbf x}^{(t)}$ above, we find
\begin{align} \label{eq:x2_dyn}
    e_3^{(t)}&=\mathbf e_2^\top\mathbf x^{(t)}\leq 
    \mathbf e_2^\top\bar{\mathbf x}^{(t)}
   \nonumber\\&=
  [g_{1}\Pi_{+,t}+g_{2}\Pi_{-,t}]e_3^{(t_k)}
  \nonumber \\&
+[g_{3}\Pi_{+,t}+g_{4}\Pi_{-,t}]e_2^{(t_k)}
\nonumber\\&
+[g_{5}(\Pi_{+,t}-1)+g_{6}(\Pi_{-,t}-1)]\delta/\beta, 
    \end{align}
    where we have defined $g_{1}$-$g_{8}$ as
    \begin{flalign}\label{eq:g1}
        g_{1}\triangleq
        &[\mathbf U]_{2,1}[\mathbf U^{-1}]_{1,2}
        =
        \frac{1}{2}(1-\frac{1}{\sqrt{8\omega+1}})\geq 0,&
    \end{flalign}
    \begin{flalign}\label{eq:g2}
        g_{2}\triangleq
        &[\mathbf U]_{2,2}[\mathbf U^{-1}]_{2,2}
        =
        \frac{1}{2}(1+\frac{1}{\sqrt{8\omega+1}})=1-g_1\geq 0,&
    \end{flalign}
    \begin{flalign}\label{eq:g3}
        g_{3}\triangleq
        &[\mathbf U]_{2,1}[\mathbf U^{-1}]_{1,1}
         =\frac{1}{\sqrt{8\omega+1}}\in [1/3,1],&
    \end{flalign}
    \begin{flalign}
        g_{4}\triangleq
        &[\mathbf U]_{2,2}[\mathbf U^{-1}]_{2,1}
        =-g_3,&
    \end{flalign}
    \begin{flalign}\label{eq:g5}
        g_{5}
        \triangleq
        &[\mathbf U]_{2,1}[\mathbf D^{-1}\mathbf U^{-1}]_{1,1}
        =
        \frac{1}{\lambda_+\sqrt{1+8\omega}}\geq 0,&
    \end{flalign}
    \begin{flalign}\label{eq:g6}
        g_{6}\triangleq
        &[\mathbf U]_{2,2}[\mathbf D^{-1}\mathbf U^{-1}]_{2,1}
        =
        \frac{1}{-\lambda_-\sqrt{1+8\omega}}\geq0.&
    \end{flalign}
Since $\Pi_{+,t}-\Pi_{-,t}\leq 2[\Pi_{+,t}-1]$ and $g_4=-g_3$, we can further upper bound $e_3$ as
\begin{align} \label{eq:E_2}
    &e_3^{(t)}\leq
     [g_{1}\Pi_{+,t}+g_{2}\Pi_{-,t}]e_3^{(t_k)}
      \nonumber \\&
    +2g_3[\Pi_{+,t}-1]e_2^{(t_k)}
    \nonumber\\&
    +[g_{5}(\Pi_{+,t}-1)+g_{6}(\Pi_{-,t}-1)]\delta/\beta.
\end{align}
Next, we use these results to bound $e_2$ and $e_3$ at global synchronization.
\subsection{Obtaining the upper bound of $e_2^{(t_{k+1})}$ at global synchronization.} 
To obtain the behavior of $\bar{\mathbf v}_c^{(t_{k+1})}-\bar{\mathbf v}^{(t_{k+1})}$ after global synchronization, we use the definition of $\bar{\mathbf v}_c^{(t)}$, $\bar{\mathbf v}^{(t)}$ and the global synchronization scheme in~\eqref{eq:aggr_alpha}, we have  
\begin{align}
    \bar{\mathbf v}_c^{(t_{k+1})}-\bar{\mathbf v}^{(t_{k+1})}
    =& (1-\alpha)(\bar{\mathbf v}^{(t_{k+1}-\Delta)}-\bar{\mathbf v}^{(t_{k+1}-\Delta)})
    +\alpha\left(\widetilde{\mathbf v}_c^{(t_{k+1})}-\widetilde{\mathbf v}^{(t_{k+1})}\right)
    \nonumber \\
    =&
    \alpha\left(\widetilde{\mathbf v}_c^{(t_{k+1})}-\widetilde{\mathbf v}^{(t_{k+1})}\right),
\end{align}
where $\widetilde{\mathbf v}^{(t_{k+1})}$ is the global noise-free variable right before global synchronization, as opposed to $\mathbf v^{(t_{k+1})}$ defined right after global synchronization.
Taking the norm of both hand sides of the above equality and adding over $\sum_c\rho_c$ yields
\begin{align} \label{eq:wc-w*_sync2}
        &e_2^{(t_{k+1})}=\sum_c\rho_c\Vert\bar{\mathbf v}_c^{(t_{k+1})}-\bar{\mathbf v}^{(t_{k+1})}\Vert
        = \alpha\sum_c\rho_c\left\Vert\widetilde{\mathbf v}_c^{(t_{k+1})}-
        \widetilde{\mathbf v}^{(t_{k+1})}\right\Vert.
\end{align} 
Note that, since $\widetilde{\mathbf v}_c^{(t_{k+1})}$ represents the noise-free variable right before the global synchronization, the right hand side above can be bounded
via \eqref{eq:E_1}, yielding the final result \eqref{eq:x1_syn_inter1}.
\subsection{Obtaining upper bound of $e_3^{(t_{k+1})}$ at global synchronization.}
To obtain the behavior of $\bar{\mathbf v}^{(t_{k+1})}-\mathbf w^*$ after global synchronization,  we use the definition of $\bar{\mathbf v}^{(t)}$ and the global synchronization scheme in~\eqref{eq:aggr_alpha}, we have  
\begin{align}
    & \bar{\mathbf v}^{(t_{k+1})}-\mathbf w^*
        = (1-\alpha)[\bar{\mathbf v}^{(t_{k+1}-\Delta)}-\mathbf w^*]
        +\alpha\left[\widetilde{\mathbf v}^{(t_{k+1})}-\mathbf w^*\right].
\end{align} 
Note that $\widetilde{\mathbf v}^{(t)}$ is equivalent to $\bar{\mathbf v}^{(t)}$ before conducting global synchronization. Taking the norm of both hand sides of the above equality and applying the triangle inequality
gives us
\begin{align} \label{eq:wc-w*_sync_tmp}
        &e_3^{(t_{k+1})}= \Vert\bar{\mathbf v}^{(t_{k+1})}-\mathbf w^*\Vert
        \leq (1-\alpha)\Vert\bar{\mathbf v}^{(t_{k+1}-\Delta)}-\mathbf w^*\Vert
        +\alpha\Vert\widetilde{\mathbf v}^{(t_{k+1})}-\mathbf w^*\Vert.
\end{align} 
We further upper bound the right hand using~\eqref{eq:E_2} to obtain the result in\eqref{eq:wc-w*_sync1}.
\end{proof}

\begin{fact}\label{fact:1} 
Consider $n$ random real-valued vectors $\mathbf x_1,\cdots,\mathbf x_n\in\mathbb R^m$, the following inequality holds: 
 \begin{equation}
     \sqrt{\mathbb E\left[\Big\Vert\sum\limits_{i=1}^{n} \mathbf x_i\Big\Vert^2\right]}\leq \sum\limits_{i=1}^{n} \sqrt{\mathbb E[\Vert\mathbf x_i\Vert^2]}.
 \end{equation}
\end{fact}
\begin{proof} Note that
    \begin{align}
        &\sqrt{\mathbb E\left[\Big\Vert\sum\limits_{i=1}^{n}\mathbf x_i\Big\Vert^2\right]}
        =
        \sqrt{\sum\limits_{i,j=1}^{n}\mathbb E [\mathbf x_i^\top\mathbf x_j]}
        \overset{(a)}{\leq}
\sum\limits_{i,j=1}^{n}\sqrt{\mathbb E [\Vert\mathbf x_i\Vert^2] \mathbb E[\Vert\mathbf x_j\Vert^2]]}
        =
        \sum\limits_{i=1}^{n} \sqrt{\mathbb E[\Vert\mathbf x_i\Vert^2]},
    \end{align}
    where $(a)$ follows from Holder's inequality, $\mathbb E[|XY|] \leq \sqrt{\mathbb E[|X|^2]\mathbb E[ |Y|^2]}$.
\end{proof}

\begin{fact}\label{fact:3} Let $f(\cdot)$ be $\mu$-strong convex and $\beta$-smooth and $\eta\leq\frac{2}{\beta+\mu}$, the following inequality holds
\begin{align}
    \left\Vert\mathbf w_1-\mathbf w_2 -\eta(\nabla f(\mathbf w_1)-\nabla f(\mathbf w_2))\right\Vert
    \leq (1-\mu\eta)\left\Vert\mathbf w_1-\mathbf w_2\right\Vert,\ \forall \mathbf w_1,\mathbf w_2\in\mathbb R^M.
\end{align}    
\end{fact}
\begin{proof}
    \begin{align} \label{eq:stx_3}
        &\left\Vert\mathbf w_1-\mathbf w_2 -\eta(\nabla f(\mathbf w_1)-\nabla f(\mathbf w_2))\right\Vert
        \nonumber \\&
        =
        \sqrt{\Vert\mathbf w_1-\mathbf w_2\Vert^2+\eta^2\Vert\nabla f(\mathbf w_1)-\nabla f(\mathbf w_2)\Vert^2-2\eta(\mathbf w_1-\mathbf w_2)^\top(\nabla f(\mathbf w_1)-\nabla f(\mathbf w_2))}
        \nonumber \\&
        \overset{(a)}{\leq} 
        \sqrt{\left(1-\eta\frac{2\mu\beta}{\mu+\beta}\right)\Vert\mathbf w_1-\mathbf w_2\Vert^2-\eta\left(\frac{2}{\mu+\beta}-\eta\right)\Vert\nabla f(\mathbf w_1)-\nabla f(\mathbf w_2)\Vert^2}
        \nonumber \\&
        \overset{(b)}{\leq}
        (1-\eta\mu)\Vert\mathbf w_1-\mathbf w_2\Vert,
    \end{align}
    where $(a)$ comes from~\cite[Theorem 2.1.12]{Nesterov}, i.e., $(\mathbf w_1-\mathbf w_2)^\top(\nabla f(\mathbf w_1)-\nabla f(\mathbf w_2))\geq\frac{\mu\beta}{\mu+\beta}\Vert\mathbf w_1-\mathbf w_2\Vert^2+\frac{1}{\mu+\beta}\Vert\nabla f(\mathbf w_1)-\nabla f(\mathbf w_2)\Vert^2$ and $(b)$ results from
    $\Vert\nabla f(\mathbf w_1)-\nabla f(\mathbf w_2)\Vert\geq \mu\Vert\mathbf w_1-\mathbf w_2\Vert$ (strong convexity)
 and $\eta\leq \frac{2}{\mu+\beta}$.
\end{proof}


\pagebreak

\end{document}